\documentclass[10pt, conference]{IEEEtran}

\IEEEoverridecommandlockouts                              

\input{bib.def.short}

\author{Constantin Selzer$^{1}$ and Fabian B. Flohr$^{1}$
\thanks{$^{1}$ Department of Electrical Engineering and Information Technology, Intelligent Vehicles Lab (IVL), Munich University of Applied Science, Lothstraße 34, 80335 Munich, German {\tt\small Constantin.Selzer@hm.edu}}%
}
\usepackage{cite}
\usepackage{amsmath,amssymb,amsfonts}
\usepackage{algorithmic}
\usepackage{graphicx}
\usepackage{tabularx}
\usepackage{textcomp}
\usepackage{adjustbox}
\usepackage{subcaption}
\usepackage{xcolor}
\usepackage{hyperref}
\def\BibTeX{{\rm B\kern-.05em{\sc i\kern-.025em b}\kern-.08em
    T\kern-.1667em\lower.7ex\hbox{E}\kern-.125emX}}
\begin{document}

\newcommand{\TODO}[1]{{\textcolor{red}{{#1}}}} 
\newcommand{\mycomment}[1]{}

\title{DeepUrban: Interaction-aware Trajectory Prediction and Planning for Automated Driving by Aerial Imagery\\}

\IEEEoverridecommandlockouts
\IEEEpubid{\makebox[\columnwidth]{979-8-3315-0592-9/24/\$31.00~\copyright2024 IEEE \hfill}\hspace{\columnsep}\makebox[\columnwidth]{ }}

\maketitle

\IEEEpubidadjcol

\begin{abstract}

The efficacy of autonomous driving systems hinges critically on robust prediction and planning capabilities. However, current benchmarks are impeded by a notable scarcity of scenarios featuring dense traffic, which is essential for understanding and modeling complex interactions among road users. To address this gap, we collaborated with our industrial partner, DeepScenario, to develop "DeepUrban"—a new drone dataset designed to enhance trajectory prediction and planning benchmarks focusing on dense urban settings. DeepUrban provides a rich collection of 3D traffic objects, extracted from high-resolution images captured over urban intersections at approximately 100 meters altitude. The dataset is further enriched with comprehensive map and scene information to support advanced modeling and simulation tasks. We evaluate state-of-the-art (SOTA) prediction and planning methods, and conducted experiments on generalization capabilities. Our findings demonstrate that adding DeepUrban to nuScenes can boost the accuracy of vehicle predictions and planning, achieving improvements up to 44.1\% / 44.3\% on the ADE / FDE metrics. Website: \href{https://iv.ee.hm.edu/deepurban}{https://iv.ee.hm.edu/deepurban}

\end{abstract}


\section{Introduction}
%
Prediction and planning benchmarks are integral to the advancement of autonomous driving systems, serving as critical tools for assessing the efficacy of various approaches and facilitating their comparative analysis. Over recent years, the research community has increasingly turned its attention to complex urban scenarios, leading to the development and publication of numerous autonomous driving datasets designed to address these challenges. Notable examples include CommonRoad (2017)~\cite{CommonRoad2017}, nuScenes (2019)~\cite{nuScenes2020}, Interaction (2019)~\cite{Interaction2019}, nuPlan (2021)~\cite{NuPlan2021}, Waymo Open Motion (2021)~\cite{WAYMO2021}, and the Argoverse series (2019, 2021)~\cite{Argoverse22023}~\cite{Argoverse2019}.
\begin{figure}[ht!]
    \centering
    \includegraphics[width=1\linewidth]{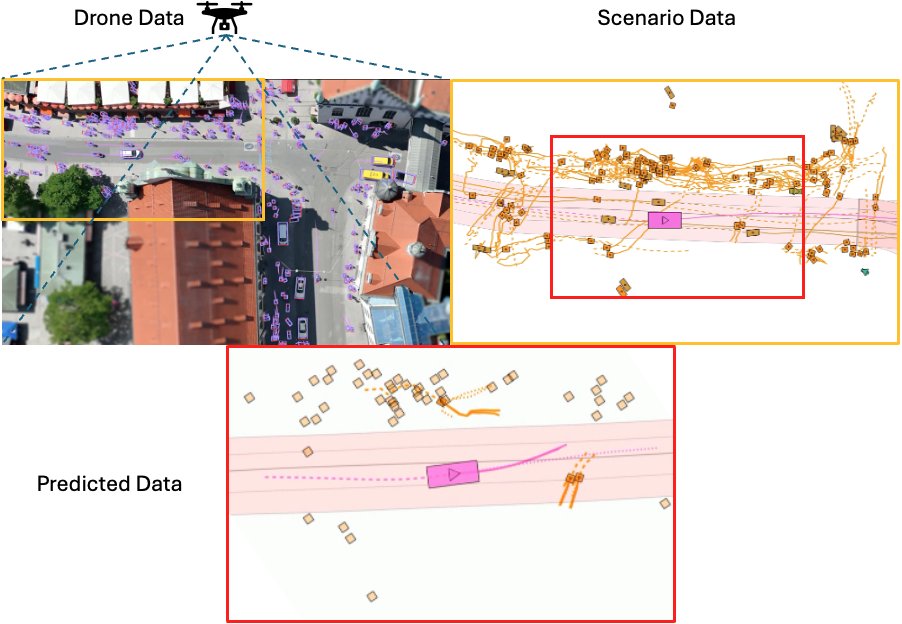}
    \caption{DeepUrban dataset: Data pipeline showing detected and tracked agents including 14 different agent types (top left), generated scenario data from raw inputs (top right) and predicted / planned clique agents using the ScePT algorithm~\cite{ScePT2022} (bottom).}
    \label{fig:title}
\end{figure}
While the diversity of data in existing autonomous driving datasets is commendable, there remains a notable gap in capturing high-density interactions among road users. This limitation is critical, as it affects the ability of autonomous systems to effectively navigate and respond in densely populated urban environments.

One reason for this gap is that the majority of available datasets predominantly focus on environments in America and Asia, where road densities are approximately 36.02~km~/~100~km² and 68~km~/~100~km², respectively. This is in contrast to Germany, which features one of the highest road densities globally, approximately 180.51~km~/~100~km². Even in rural areas interaction between traffic participants is more pronounced, and visible.
%
And in contrast to many other parts of the world, European cities exhibit a smooth integration between different types of traffic participants, such as vehicles, and vulnerable road users (VRUs). This integration is facilitated also by traffic rules, and infrastructure that are tailored to enhance cooperation, and safety among all road users.

This discrepancy highlights a significant variation in the environmental complexity that datasets cover, suggesting an opportunity to expand dataset diversity to include regions with higher road densities like Germany. Such an expansion could potentially enhance the robustness, and applicability of autonomous driving technologies, ensuring their adaptability to a wider range of global, more complex traffic conditions, and infrastructural densities.

In collaboration with our industrial partner, DeepScenario, we created the "DeepUrban (V1)"—dataset to enhance benchmarking of trajectory prediction, and planning methods with a focus on dense urban settings with a heavy interaction between road users. DeepUrban is distinguished by its rich collection of relevant 3D traffic objects, extracted from high-resolution images captured by drones (cf. Figure~\ref{fig:title}). The dataset is further enriched with comprehensive map, and scene information to support advanced modeling, and simulation tasks.

In extensive experiments, we showcase results of a state-of-the-art method for a data-driven, and scene-consistent prediction, and planning (ScePT~\cite{ScePT2022}). We explore the generalization capabilities across areas, and other datasets.
While comparing different approaches on various datasets, and data formats can be challenging, we have integrated our new dataset into the TrajData dataloader~\cite{TrajData2023}, which already supports numerous other datasets, and unifies the access to various data sources. 

Through our partner DeepScenario we provide access to the recorded, and processed aerial data. Furthermore, we provide the full data pipeline for preprocessing the data for prediction, and planning tasks which is illustrated in Figure~\ref{fig:title}.
In the future, we plan to further extend the dataset, and facilitate online-benchmarking on a hidden test set. 

In summary, our contributions are two-fold: 
Together with our industrial Partner DeepScenario we provide a rich dataset recorded from on-board a drone in various dense urban locations. We extract over 5604 high-density urban traffic scenarios with heavy VRU interactions between road users. Further, we integrate the DeepUrban data into the TrajData dataloader~\cite{TrajData2023} to facilitate easier benchmarking, also across datasets.

We then highlight the generalization capabilities of the DeepUrban dataset by benchmarking state-of-the-art trajectory prediction, and planning architectures \cite{ScePT2022}. Cross-dataset experiments suggest an improvement in accuracy when expanding existing datasets with the DeepUrban dataset. By expanding the nuScenes training dataset with our scenarios from 'Munich Tal', we show an improvement of 44.1\%,44.3\%, and 49.6\% in ADE, FDE, and collision score respectively for vehicles.



\section{Related Work}
In the following we review recent works for datasets, and methods on trajectory prediction and planning.

\textbf{Trajectory Prediction and Planning Datasets}:
Public prediction, and planning datasets are instrumental in advancing autonomous driving technologies, providing a foundation for developing, testing, and enhancing vehicle systems. 

Automotive datasets like Argoverse~\cite{Argoverse22023, Argoverse2019}, Waymo Open Motion~\cite{WAYMO2021}, and nuScenes~\cite{nuScenes2020}, nuPlan~\cite{NuPlan2021}, Lyft Level 5~\cite{Lyft2021}  are recorded from on-board a moving vehicle, and provide comprehensive coverage across several US cities (Argoverse, Waymo Open Motion, nuScenes), and Singapore (nuScenes), capturing a high diversity in road user dynamics, and interactions between different agent types, thereby facilitating studies in urban settings. 

European datasets like CommonRoad, and KITTI provide varied scenarios from Germany, ranging from lane following in rural environments, and highways to urban intersections, thereby enhancing perception, and planning capabilities in non-US environments \cite{CommonRoad2017, KITTI2013}.

The Interaction~\cite{Interaction2019} dataset provides 16.5 hours of trajectory data in four different countries. The data covers mainly highly interactive scenarios such as merging, and roundabout scenes.

The ETH, and UCY dataset~\cite{ETH2009, UCY2007}, though not primarily for autonomous driving, offer rich data for VRU perception tasks, thus supporting the development of safety mechanisms in autonomous vehicle systems.

Using drone data (e.g. inD~\cite{inD2020}, Stanford Drone (2016)~\cite{StanfordDrone2016})for prediction, and planning provides a broader perspective compared to ego-vehicle sensors, capturing extensive areas, and different agents to better understand traffic dynamics. Drones offer unobstructed views, crucial for detecting non-line-of-sight events, and enhancing the reliability of data in dense urban environments. Their unobstructed global view particularly enables the potential use of every detected vehicle as an ego-vehicle for planning. Additionally, the drones flexibility allows for adaptable, and cheap data collection across specific areas. 

We noticed that the above-mentioned datasets often have a limited amount of scenarios with high-density environments, and complex interactions between multiple agents, leading to limited modeling capabilities of such complex behaviors. This is also confirmed by our experiments. Thus we suggest that expanding existing datasets with such high-density scenarios can improve the overall accuracy.

Table~\ref{tab:comparison} shows some statistics on available datasets. With the current dataset configuration, DeepUrban reaches the highest VRU density with 12h of recording.
\begin{table}
    \scriptsize
    \vspace{0.1cm}
    \centering
    \begin{tabularx}{\linewidth}{|X||X|X|X|X|}
        \hline
        & \textbf{Dataset Hours} & \textbf{Trajectories} & \textbf{VRUs in percentage} & \textbf{Agent types} (with Subcategories) \\
        \hline
        \textbf{Waymo} &  574h & 7.64m & 10\% & 3 \\
        \hline
        \textbf{Lyft} & 1118h & 53.4m & 3\% & 4 \\
        \hline
        \textbf{nuScenes} & 5.5h & 4.3k & 23\% & 5(23) \\
        \hline
        \textbf{Interaction} & 16.5h & 40k & 5\% & 3 \\
        \hline
        \textbf{inD} & 10h & 11.5k & 43\%& 6 \\
        \hline
        \textbf{DeepUrban} (ours) & 12h & 208.3k & 95\% & 14 \\
        \hline
    \end{tabularx}
    \caption{Comparison with existing autonomous driving datasets for prediction and planning.}
    \label{tab:comparison}
\end{table}

\textbf{Data-driven Trajectory Prediction and Planning}:
The planning model PDM (Planning-Driving Model)~\cite{PDM2023}, builds on the foundations of the IDM (Intelligent Driver Model) by incorporating an advanced ego-forecasting component. This integration introduces multiple enhancements, one of which is the use of the GC-PGP (Goal-Centric Policy Gradient Planner)~\cite{GC-PGP2023}. Another noteworthy model, Hoplan~\cite{hoplan2023}, employs a unique approach by rasterizing both current and historical trajectories of all agents onto a map, thus creating a heatmap prediction. This heatmap is then utilized by a post-solver motion planner to refine the driving trajectory for the autonomous vehicle.

In the realm of hierarchical modeling, sophisticated techniques are employed to predict and plan trajectories. GameFormer~\cite{GameFormer2023} introduces a hierarchical Transformer structure that specifically enhances interaction predictions and concentrates on the trajectory of the ego vehicle by utilizing the results from previous decoding layers. Similarly, ScePT (Scene Prediction Transformer)~\cite{ScePT2022} takes a policy planning-based approach to produce trajectories that are consistent with the scene. It achieves this by segmenting the scene into interactive groups and making conditional predictions based on the dynamics within these groups.

Another significant advancement in autonomous driving technologies is the integration of prediction, planning, and control systems. DiffStack~\cite{diffstack2022} leverages differentiable optimization techniques, which allow for the joint optimization of these components through backpropagation. This method significantly improves the efficiency and effectiveness of the autonomous driving stack.
\section{Dataset DeepUrban}
DeepUrban V1 (i.e. the dataset used in this work) is derived from drone footage captured at various locations, primarily focusing on Germany. These scenarios are particularly rich in interactions involving VRUs, featuring a total of 14 different road user types, including the increasingly prevalent e-scooters. E-scooters are notable for their unique dynamics, characterized by small size and high acceleration capabilities.

The drone data underwent several preprocessing steps to render it suitable for use in a prediction and planning benchmark. These steps are elaborated upon in subsequent sections. Additionally, the initial scenarios have been integrated into the TrajData~\cite{TrajData2023} dataloader, which is already equipped to handle numerous datasets mentioned above, and facilitates the use of different datasets in a unified manner.

\subsection{Raw Data and Autolabeling pipeline}
The used raw data for DeepUrban, provided by our industrial partner DeepScenario, spans over four  locations~\cite{DeepScenario2023}. These locations are composed of three in Germany and one in the USA. 'Munich Tal' is a three-way intersection in Germany with a nearby pedestrian area, often leading to heavy crowds and jaywalking incidents. 'Sindelfingen Breuninger' is a parking area in Germany prone to contain irregular driving behaviors. 'Stuttgart University' is another three-way intersection in Germany, but it exhibits more fluent and suburban-like driving behavior. In the USA, 'San Francisco' features a four-way intersection with heavy traffic similar to that of 'Stuttgart University'. The raw data is captured over the urban intersection at approximately 100 meters altitude, and an approximated width and length of 150 meters depending on the location dimension, and over a few hours per location.

The trajectories were extracted from raw drone recordings using DeepScenario's proprietary computer vision pipeline. DeepScenario computes a 3D reconstruction of the static environment and leverages state-of-the-art algorithms for 3D detection and tracking of all dynamic objects in the scene. The raw data comes with an OpenDRIVE~\cite{opendrive} map and can be downloaded over a joint registration with us from DeepScenario's website\footnote{\href{https://app.deepscenario.com}{https://app.deepscenario.com}}. A detailed description of the data is beyond the scope of this work, and will be disclosed by DeepScenario in a forthcoming publication.


\subsection{Scenario filtering and extraction}
Scenarios are extracted from the raw data and categorized into distinct groups with a focus on generating situations with multiple interactions involving VRUs. To standardize the data and address issues associated with varying scenario lengths and isolated frames, each scenario is fixed at a duration of 20 seconds. This parameter is derived from currently available state-of-the-art datasets such as nuScenes~\cite{nuScenes2020}, nuPlan~\cite{NuPlan2021}, and Waymo~\cite{WAYMO2021}, which also use 20 seconds for their scenarios to further allow the comparability of our benchmark. This approach also allows for the utilization of all vehicles present in the scenes as ego-vehicles, which facilitates the prediction, and planning of diverse vehicular trajectories. Vehicles that are positioned in non-drivable or parked areas are excluded, as they are typically stationary, and unlikely to influence the paths of moving vehicles.

One of the unique benefits of using drone data for these scenarios is the capacity to include multiple ego-agents within the same scene. This allows for a comprehensive examination of the scenario from various perspectives, significantly reducing the complications associated with occlusion, commonly encountered in single ego-agent scenarios.

An agent is selected as an ego-agent when it moves at least 5 meters within a 20-second window. This movement criterion is crucial to exclude static scenarios, which would result in less engaging planning activities. Furthermore, any agents depicted in only a single frame are omitted.

The data is sampled at a frequency of 12.5Hz and an overlap of at most 5 seconds between scenarios. The maximum overlap of 5 seconds was chosen to ensure clarity and distinction, enabling the use of a 20-second duration similar to Waymo's~\cite{WAYMO2021} approach while guaranteeing at least 10 seconds of unique ego movement per scenario. Additionally, the frequency of 12.5Hz was selected to align with the framerate of the drone data supplied by DeepScenario, matching the 12.5Hz frequency of the inD drone dataset. The data sampling frequency was then resampled to 10Hz to enable more straightforward comparisons with other datasets, such as nuScenes and nuPlan at 2Hz, and Waymo at 10Hz. Detailed agent information in each scenario includes ID, Type, Boundary Box, Timesteps, and State, which comprises 3D Position, 3D Velocity, and 3D Acceleration. The map data utilized includes OpenDRIVE~\cite{opendrive}, lanelet2~\cite{lanelet22018}, and VectorMap, providing a robust framework for accurate scenario mapping.

Table~\ref{tab:deepstatis} shows statistics of available locations in the DeepUrban dataset. The scenarios for each location are split into 80\% for training, 10\% for validation, and 10\% for testing.

\begin{table}[h!]
    \scriptsize
    \centering
    \begin{tabularx}{\linewidth}{|X||X|X|X|X|}
        \hline
        & \textbf{Scenarios} & \textbf{Filtered Scenarios}  & \textbf{Trajectories} & \textbf{VRU share}\\
        \hline
        \textbf{Munich Tal} & 2.480 & 632 & 187.5k & 97.4\% \\
        \hline
        \textbf{Sindelfingen Breuninger} & 284 & 215 & 6.4k & 65.7\% \\
        \hline
        \textbf{Stuttgart University} & 2.578 & 495 & 8.3k & 65.8\% \\
        \hline
        \textbf{San Francisco}  & 262 & 211 & 6.5k & 32.6\% \\
        \hline
    \end{tabularx}
    \caption{Statistics over locations in DeepUrban, with 'filtered scenarios' referring to scenarios with ego-agents moving at least 5 meters within the 20-second frame.}
    \label{tab:deepstatis}
\end{table}

The final integration of scenarios from various locations into the TrajData~\cite{TrajData2023} facilitates comparative analyses and enhances compatibility with already implemented datasets~\cite{nuScenes2020, NuPlan2021, Interaction2019, WAYMO2021, Lyft2021}. Upon initial data load, the information is cached to expedite subsequent loading processes, resulting in significantly reduced loading times. The dataloader is equipped with comprehensive data from the scenarios, including all road user types and their metadata such as height, width, and length. It is also supplied with a vector map of the area as mentioned above. TrajData incorporates multiple functionalities for agents and maps to enable the data's utility in planning and prediction tasks, accommodating different state formats, observation formats, and agent interaction distances.



\section{Benchmark}
We apply a 80\% train / 10\% test / 10\% val data split based on each available location (cf. Table~\ref{tab:deepstatis}). The results below are computed on the validation set. Test scenarios are never touched and will be used for the upcoming online benchmarking and challenges.

\subsection{Baseline methods}



For analysis purposes, we implemented ScePT~\cite{ScePT2022} as a planning algorithm.

The ScePT model employs a discrete Conditional Variational Autoencoder (CVAE) framework for predicting trajectories within dynamic settings, particularly emphasizing the interactions among agents as depicted in a spatiotemporal scene graph. In this graph, agents are represented as nodes, while the edges, which delineate interactions, are determined based on proximity, calculated using a constant velocity model. These interactions are further defined within an adjacency matrix using an Euclidean distance threshold to vary interaction strength according to proximity. To streamline the complexity inherent in this model, the scene graph is segmented into cliques through the application of the Louvain clustering algorithm. This algorithm not only considers the histories of nodes but also takes into account the spatial context, including features such as map layouts and lane details. This methodology enables ScePT to deliver precise trajectory predictions across various scenarios, incorporating multiple modes M for different agents like pedestrians and vehicles. Additionally, a model predictive controller (MPC) is employed for planning purposes. This controller integrates the ego's reference line and predictions about agent movements within its clique to effectively manage the modes M. The choice of this algorithm is underpinned by its proven robust performance on the nuScenes dataset and its capacity to effectively handle the influence of surrounding agents, particularly in densely populated settings such as Munich Tal. Here, the algorithm's effectiveness is crucial, as including all agents might skew performance assessments. Notably, the model currently only considers the interactions among named agent types such as pedestrians and vehicles, disregarding other types of agents e.g. bicycles, motorcycles, etc...

\subsection{Metrics}
The evaluation of trajectory predictions and planning includes key metrics such as Average Displacement Error (ADE) and Final Displacement Error (FDE), which measure the mean L2 distance and the L2 distance at the endpoint between the predicted trajectories and ground truth, respectively:
\[
\text{ADE}_i = \frac{1}{T} \sum_{j=1}^{T} ||\hat{x}_{ij} - x_{ij}||_2,
\]
\[
\text{FDE}_i = ||\hat{x}_{iT} - x_{iT}||_2
\]
, where $||\cdot||_2$ denotes the L2-norm, $x_{ij}$ and $y_{ij}$ denote the coordinates of the $i$-th agent in the dataset at future timestep $j$, and $T$ denotes the final future timestep. 
Additionally, the Best-of-N metric identifies the minimum $\text{ADE}_i$ and $\text{FDE}_i$ from a set of predicted trajectory modes.
Averaging overall predicted agents results in the two scalar metrics ADE and FDE. 

Furthermore, we use the Collision Score as described in \cite{ScePT2022} to evaluate safety by assessing how often predicted trajectories fall below a safety threshold, leading to potential collisions, with a lower score indicating better performance.  
\section{Experiments}
\subsection{Baseline results}
If not otherwise mentioned, we use the 'Munich Tal' location (505 train / 63 validation / 64 test scenarios) as the base location for our experiments. Further, if not otherwise mentioned, we use only one mode for the prediction. That means that metrics are computed only on a single predicted/planned trajectory.

To evaluate the scaling performance on the DeepUrban dataset we train our baseline method (ScePT) on the first 25\%~/~50\%~/~75\% of the train scenarios, and evaluate our metrics on the validation set. See Table~\ref{tab:baseline}. Results suggest a good scaling based on the data.

\begin{figure*}[ht]
    \vspace{0.1cm}
    \begin{tabular}{ccccc}
        &  &  &  &  \\
        \includegraphics[width=0.175\linewidth]{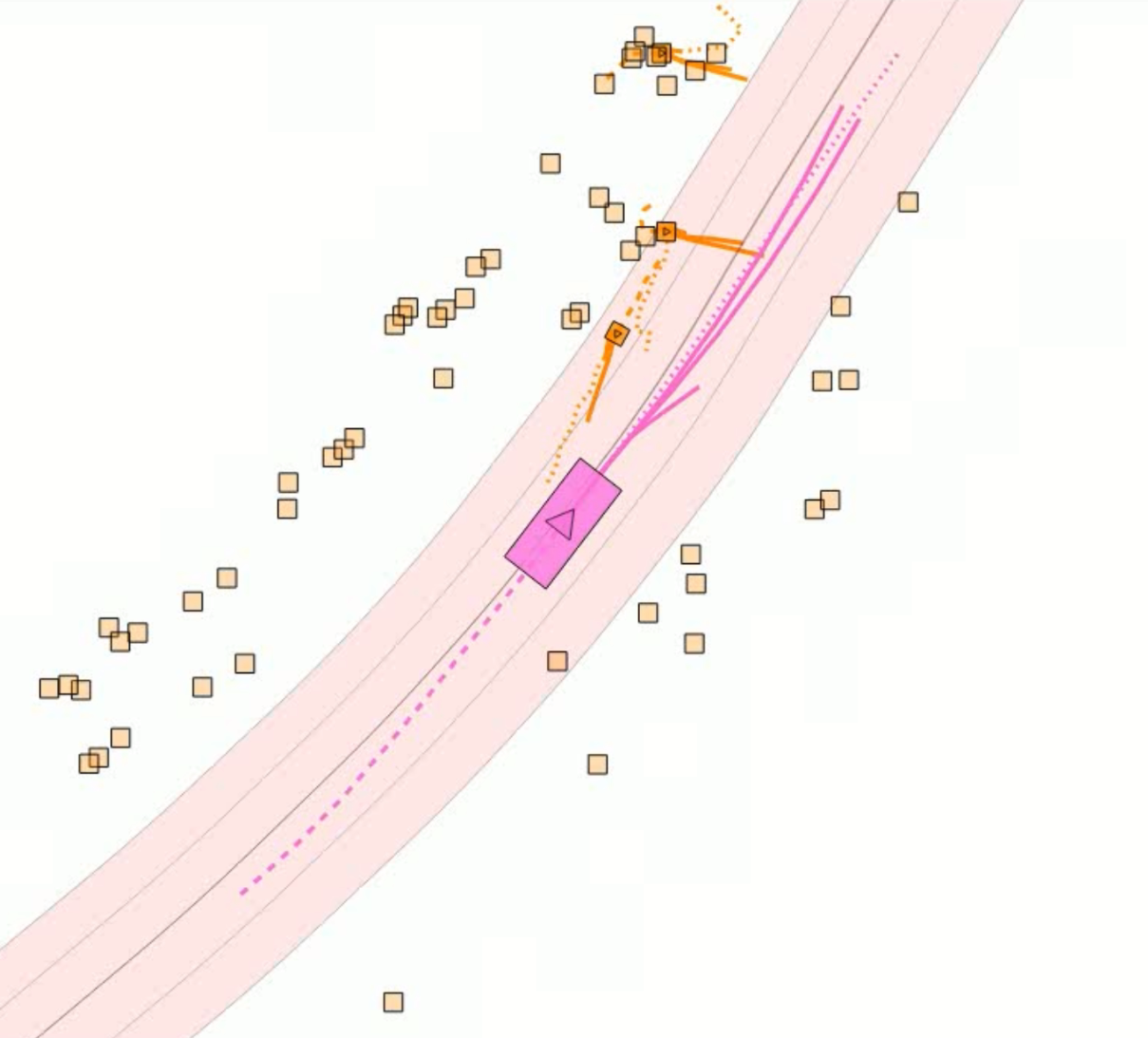} & 
        \includegraphics[width=0.175\linewidth]{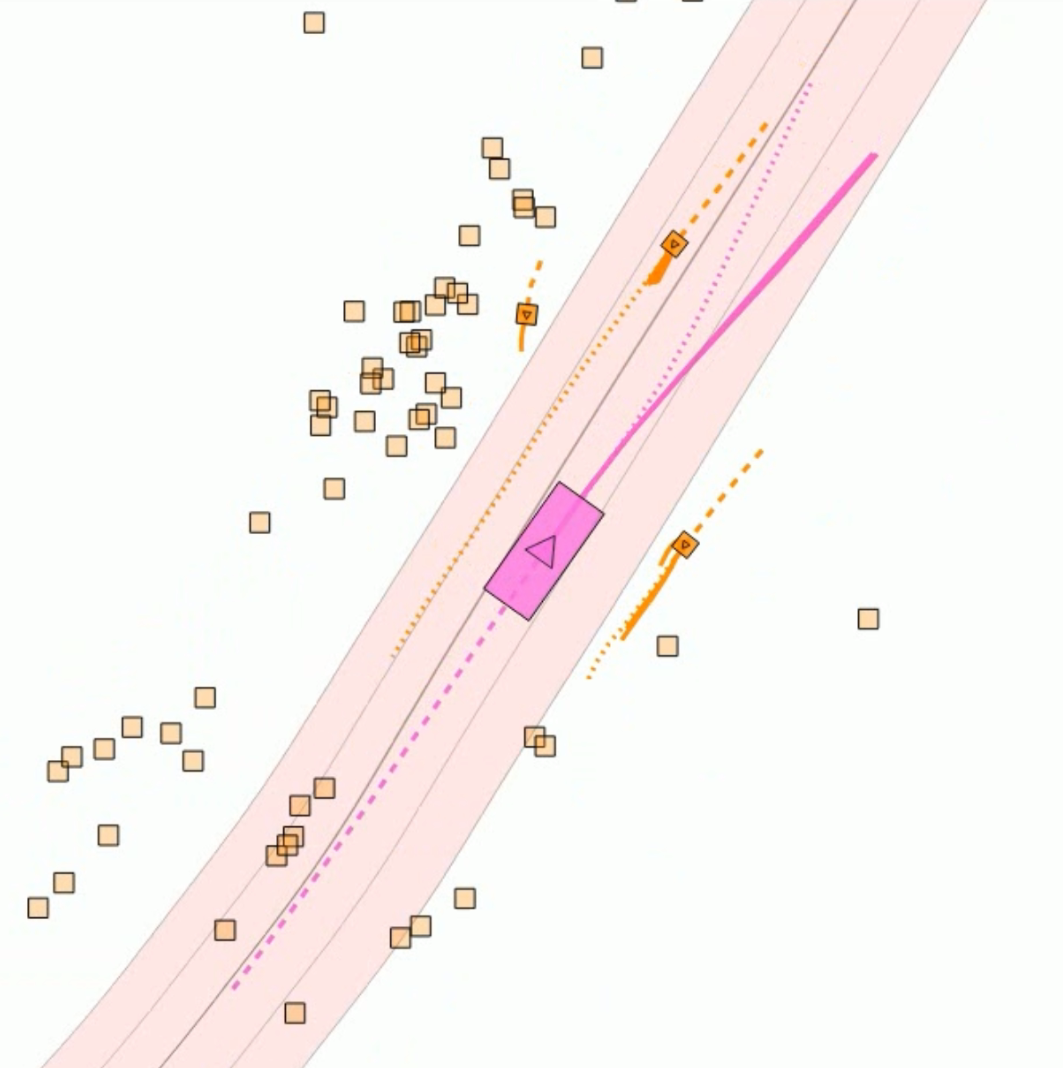} & 
        \includegraphics[width=0.175\linewidth]{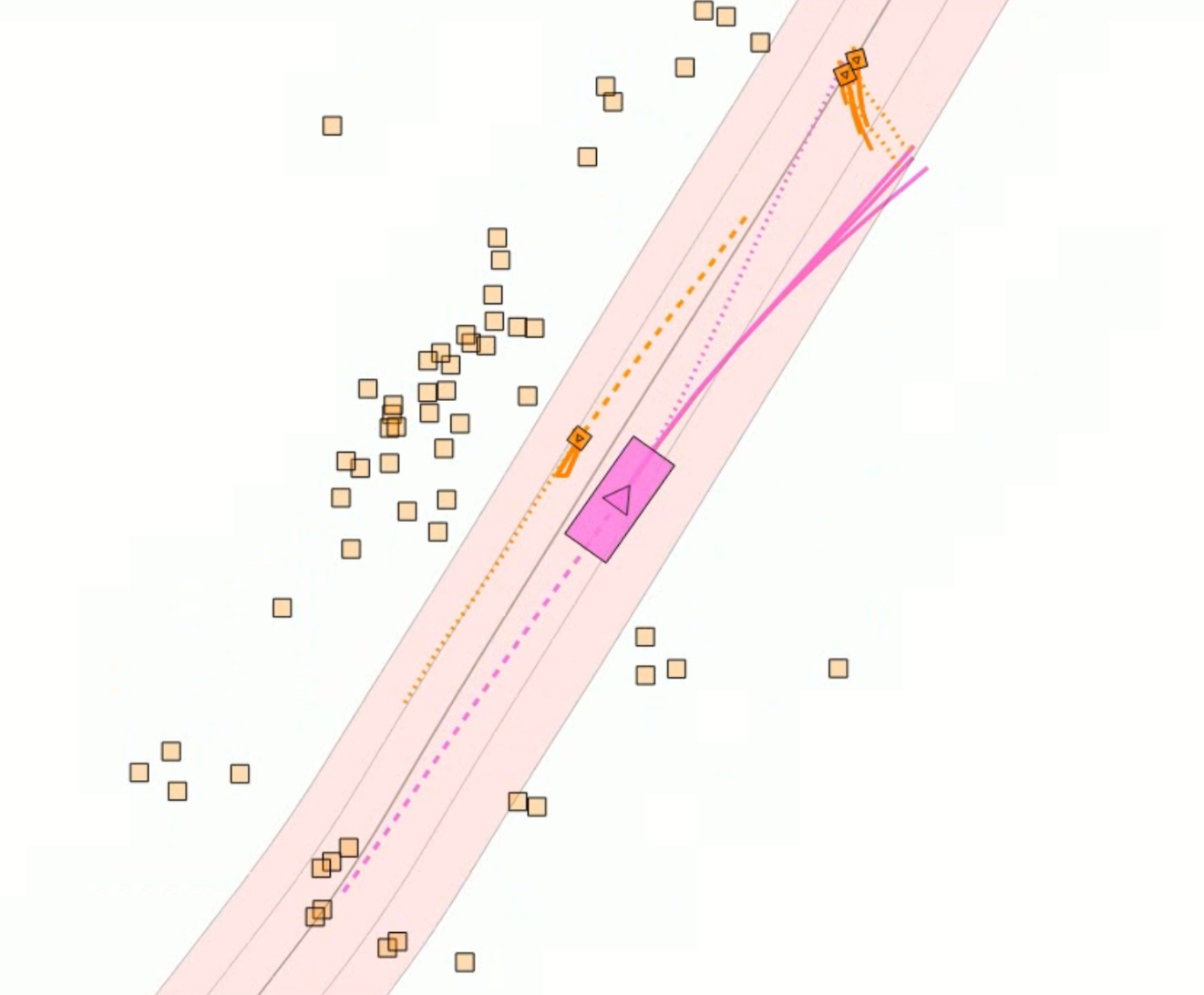} & 
        \includegraphics[width=0.175\linewidth]{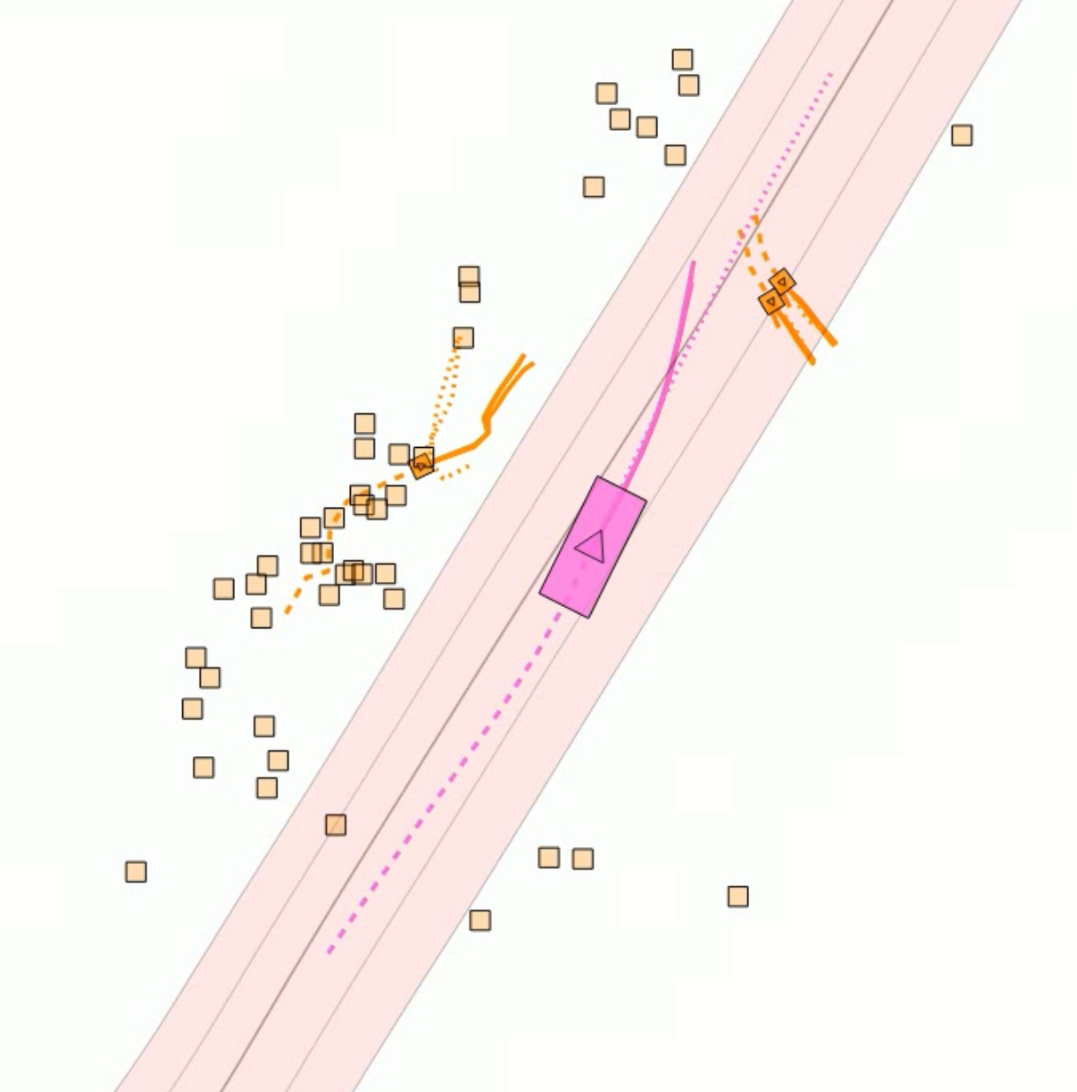} & 
        \includegraphics[width=0.175\linewidth]{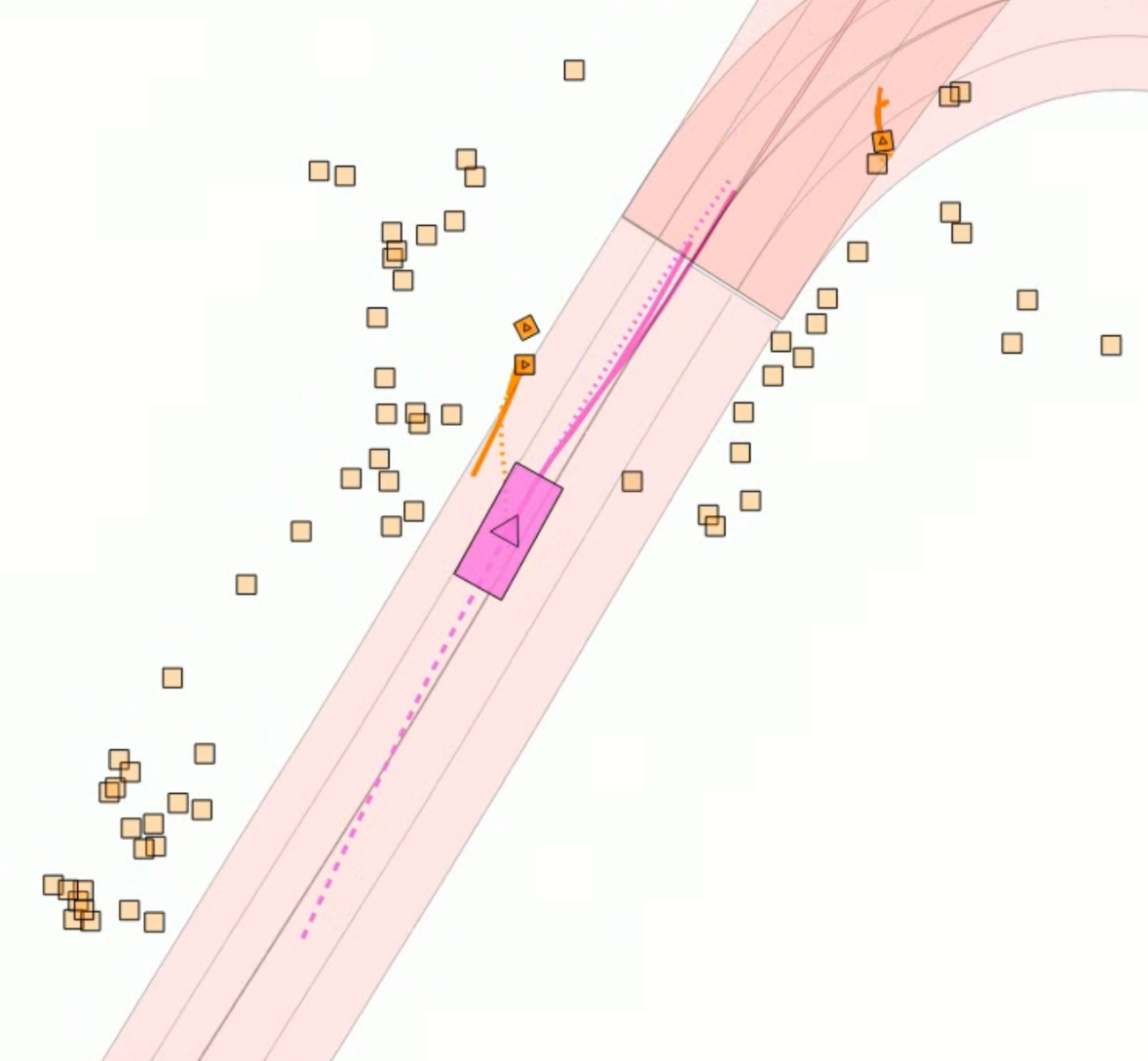} \\
        \includegraphics[width=0.175\linewidth]{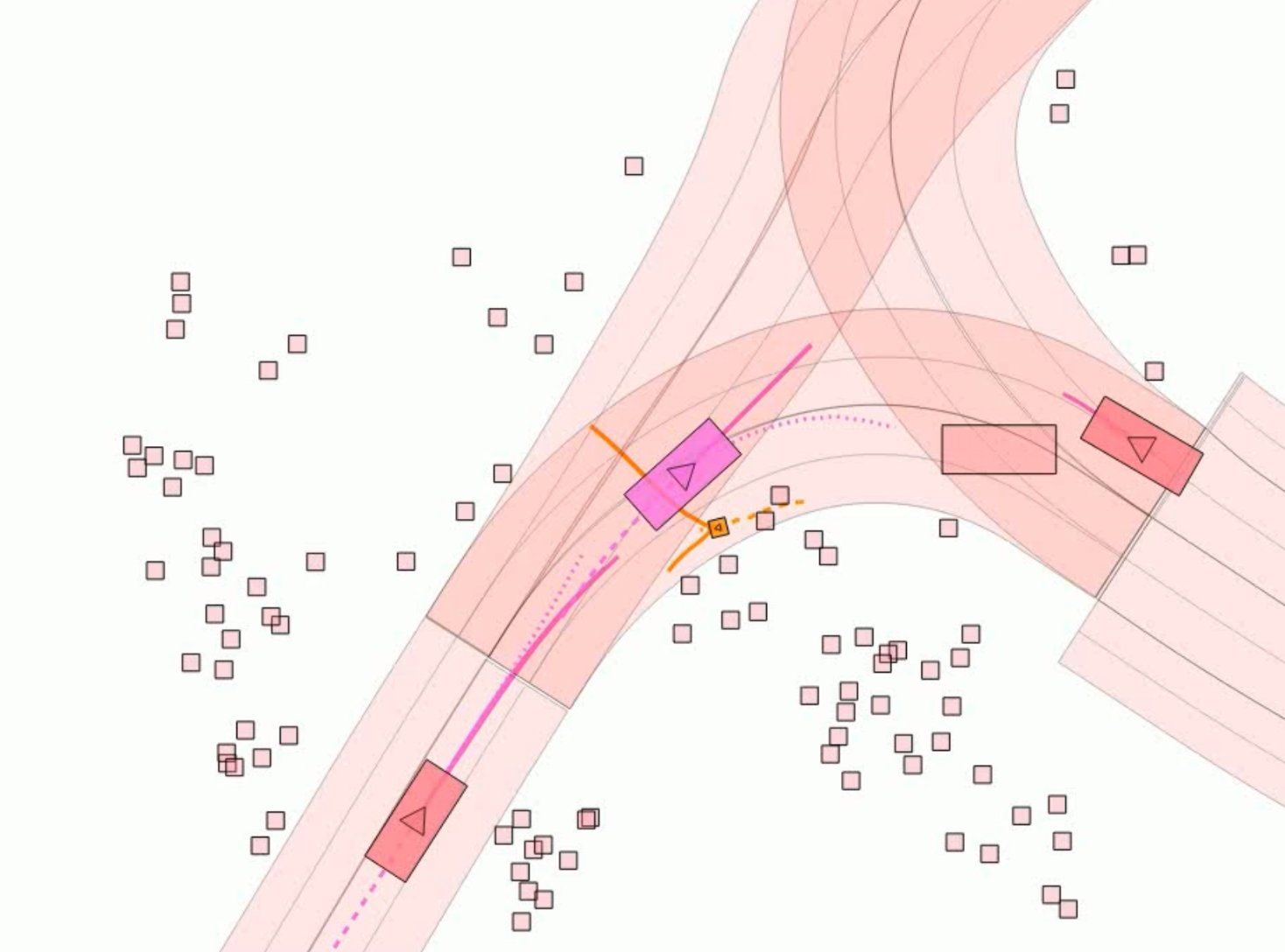} & 
        \includegraphics[width=0.175\linewidth]{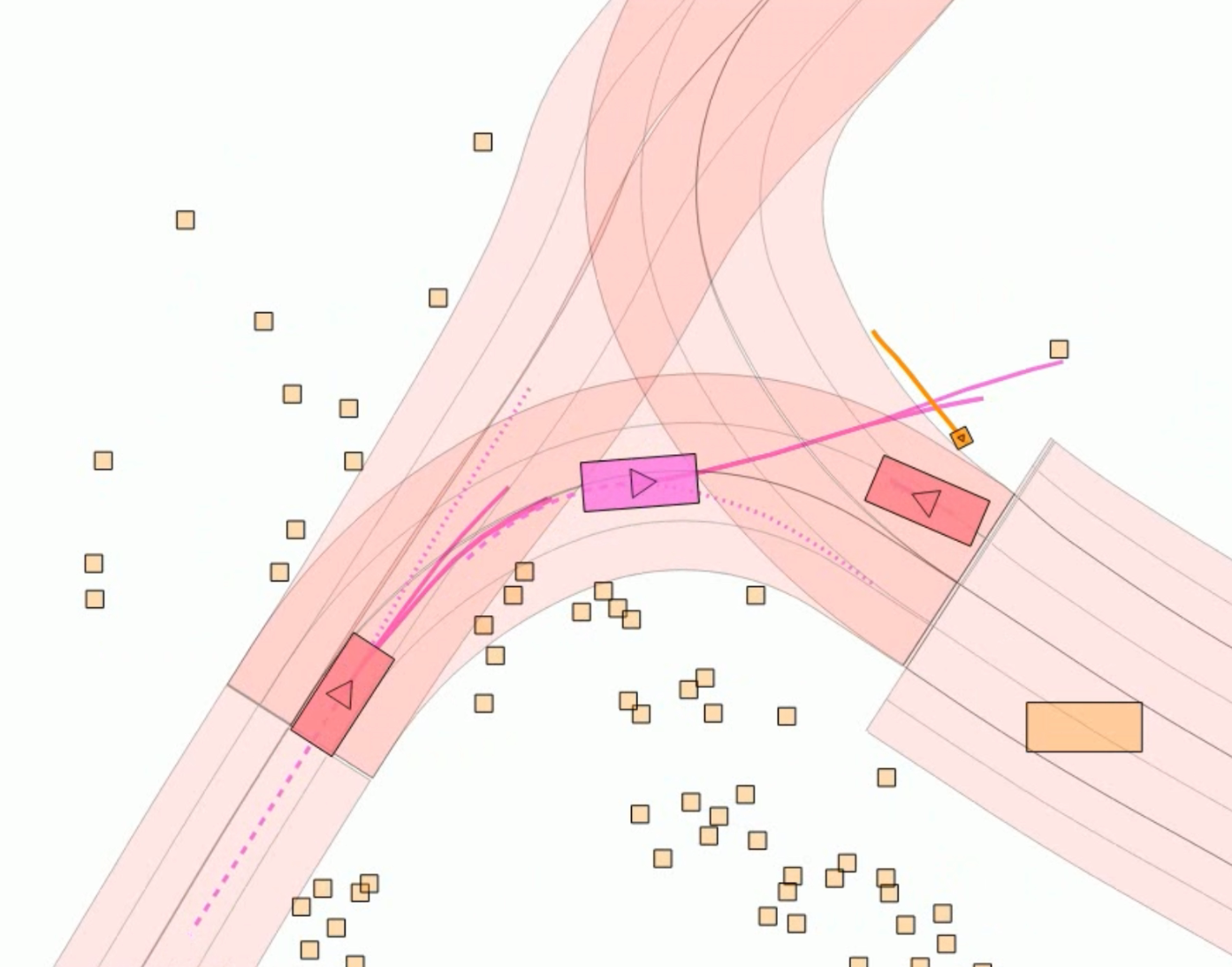} & 
        \includegraphics[width=0.175\linewidth]{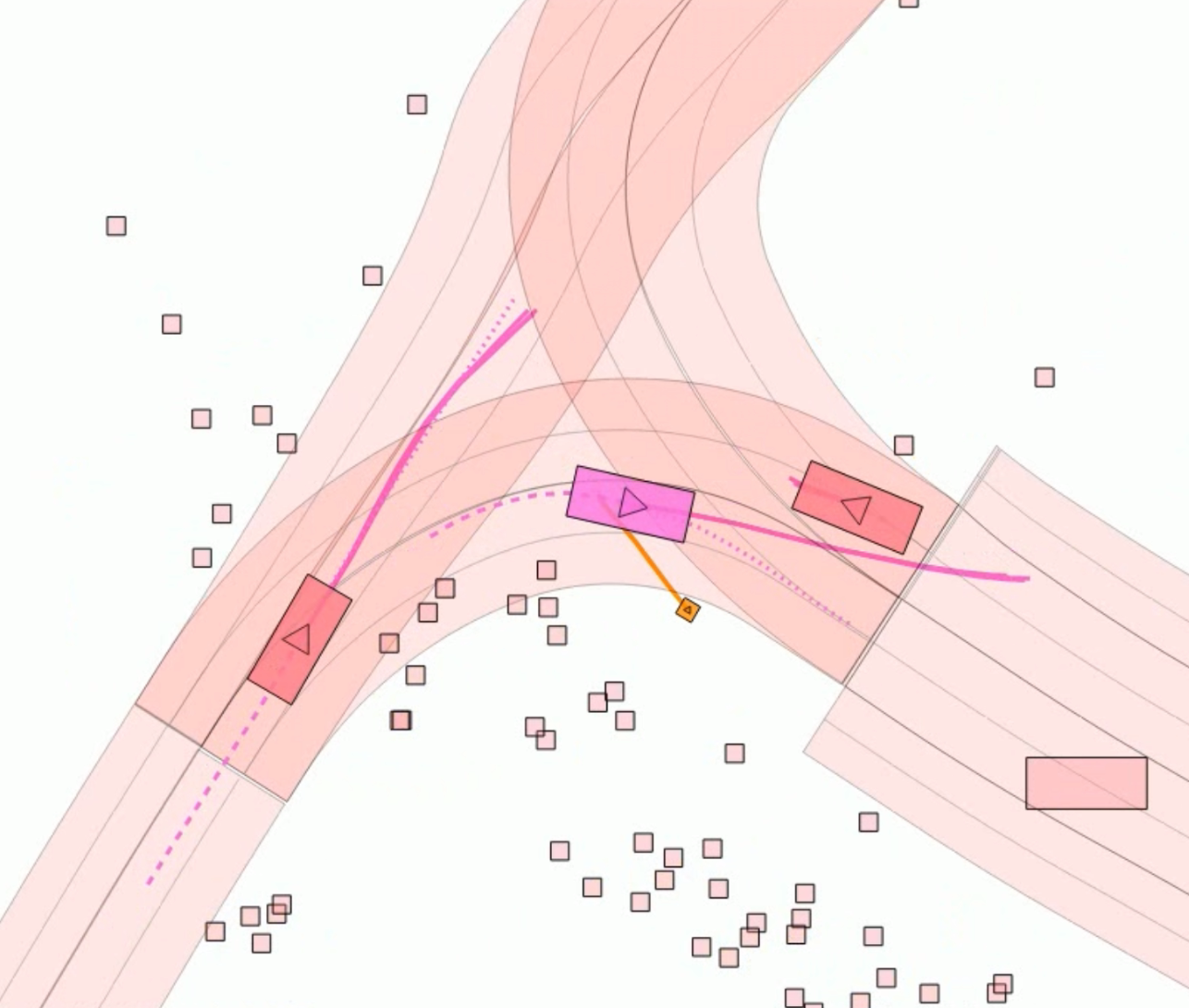} & 
        \includegraphics[width=0.175\linewidth]{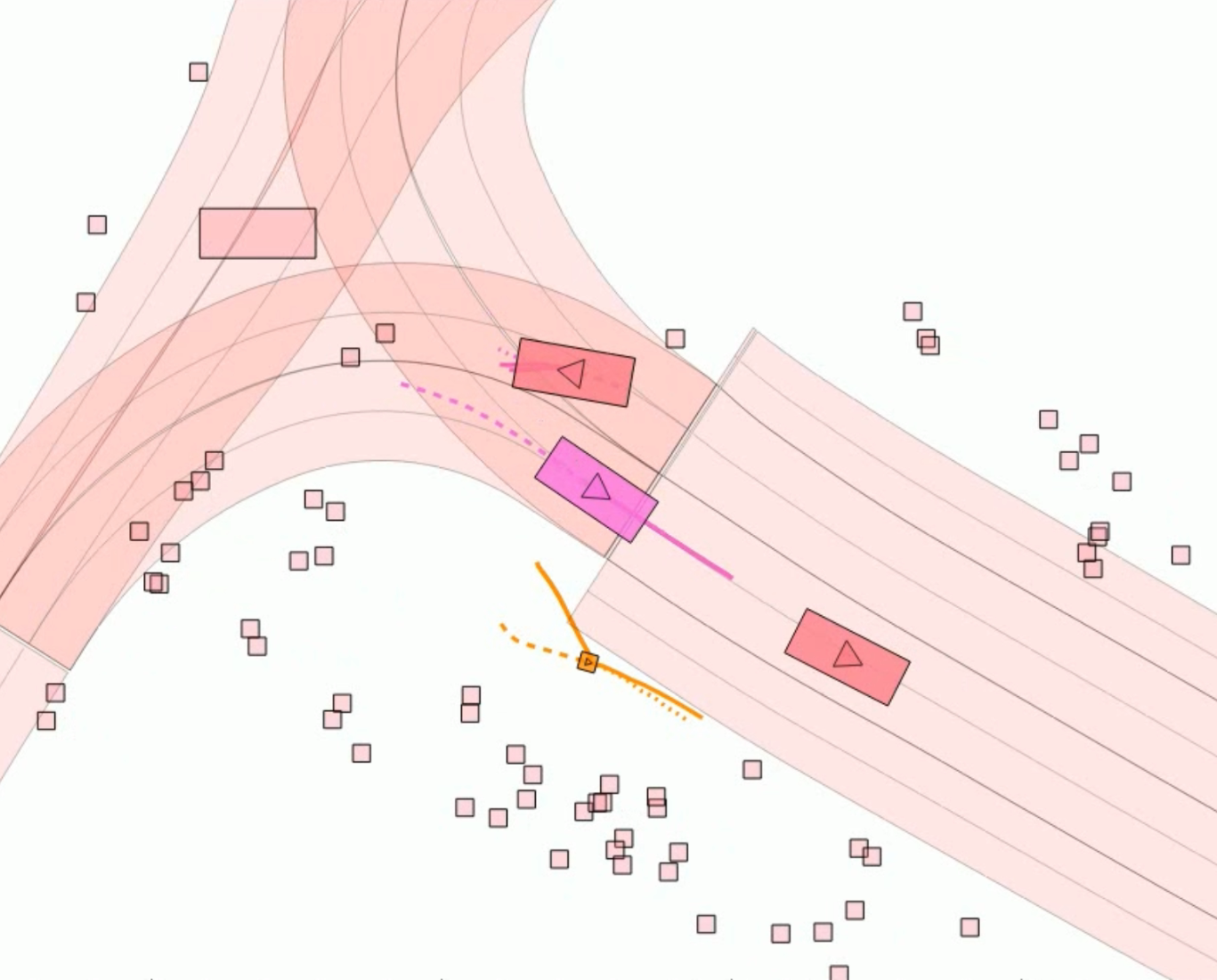} & 
        \includegraphics[width=0.175\linewidth]{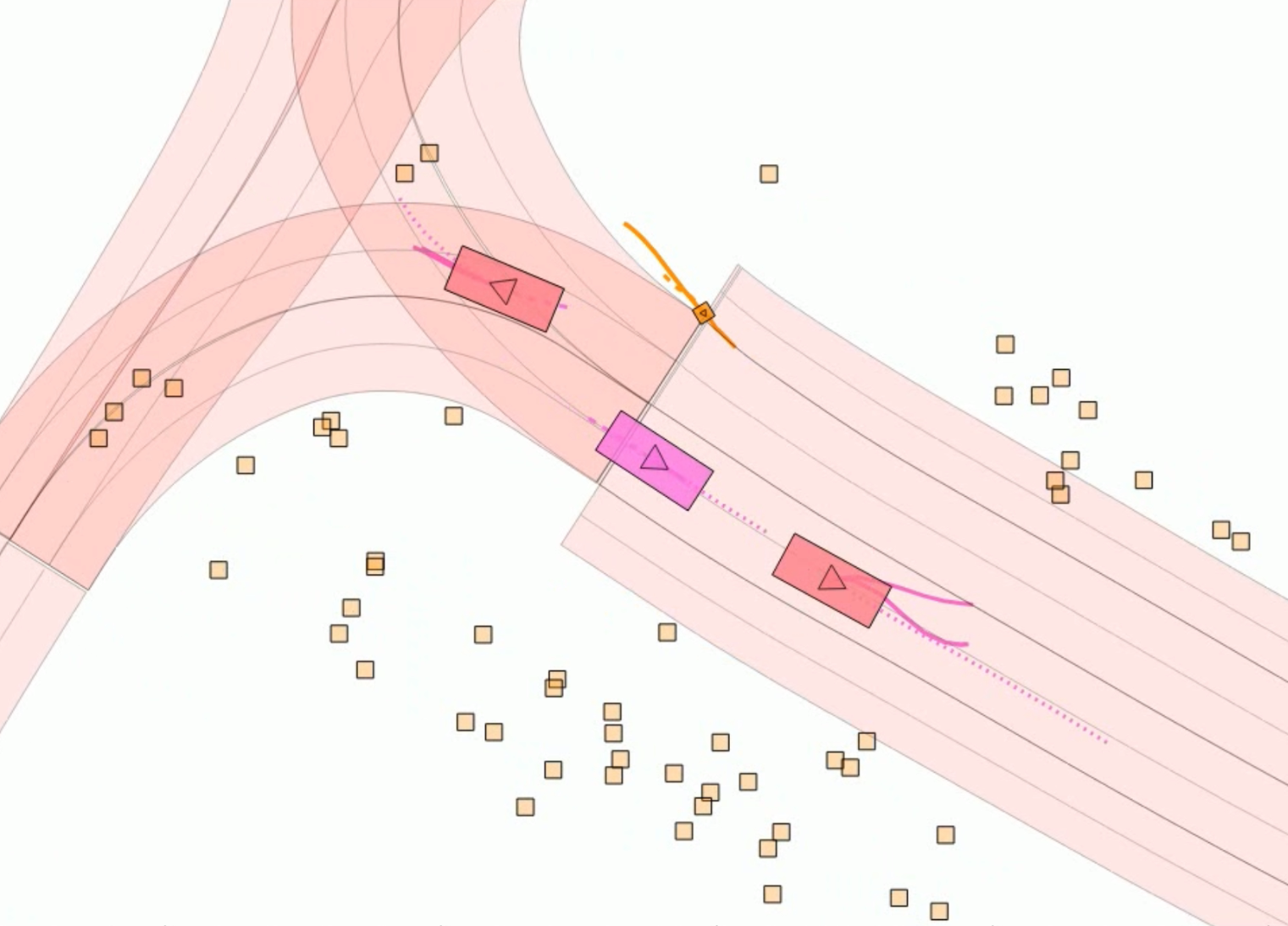} \\
        \includegraphics[width=0.175\linewidth]{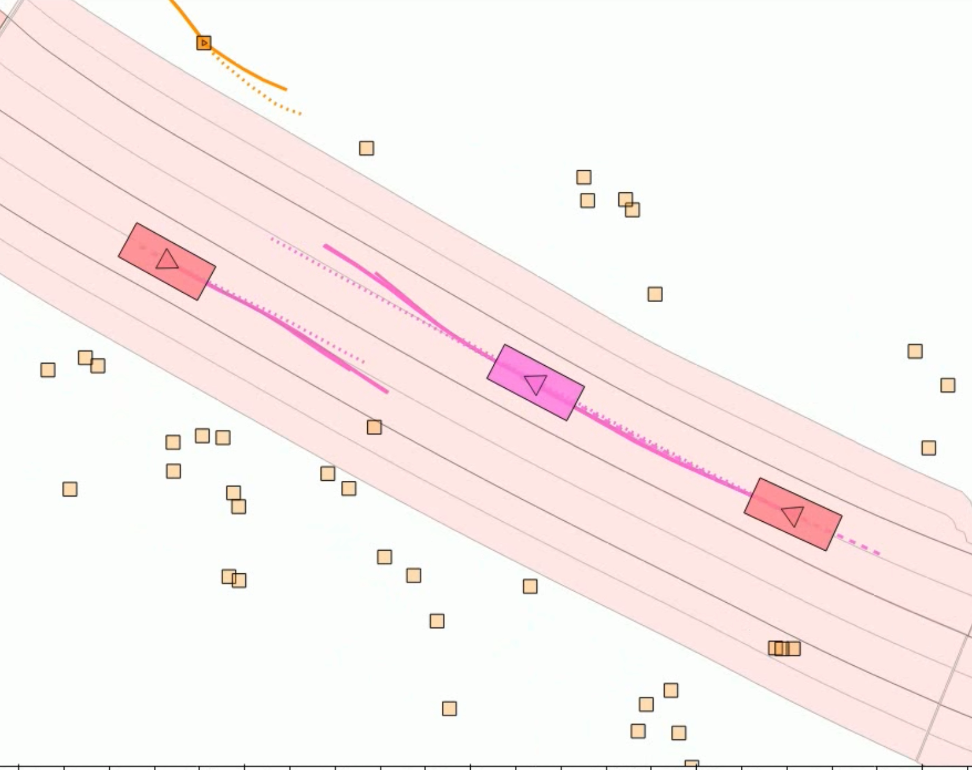} & 
        \includegraphics[width=0.175\linewidth]{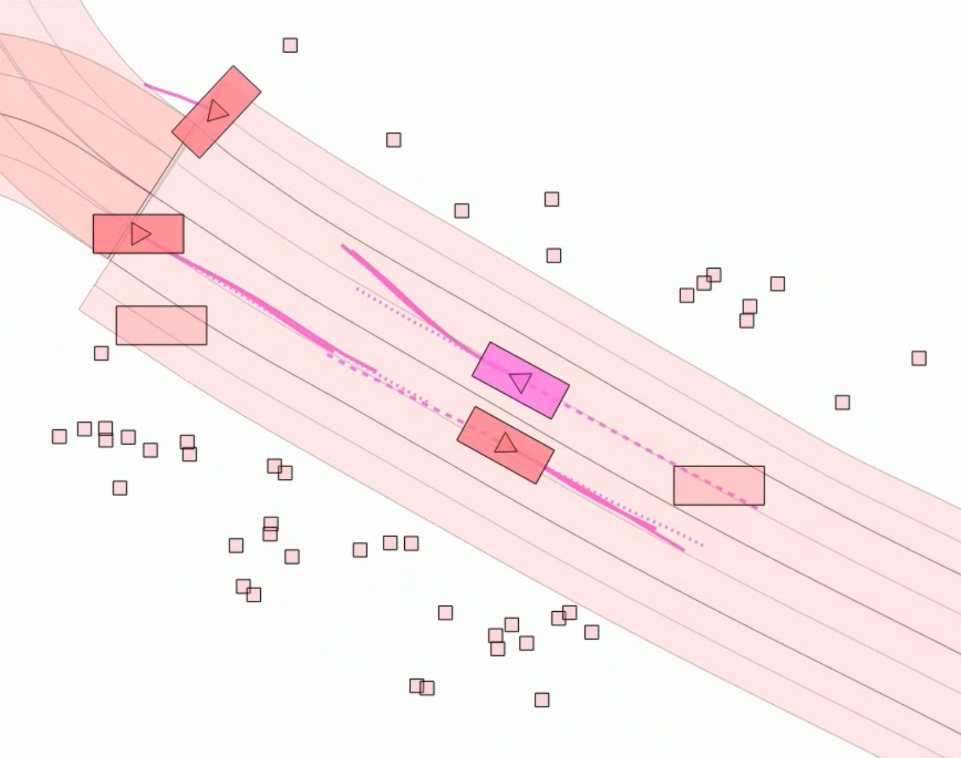} & 
        \includegraphics[width=0.175\linewidth]{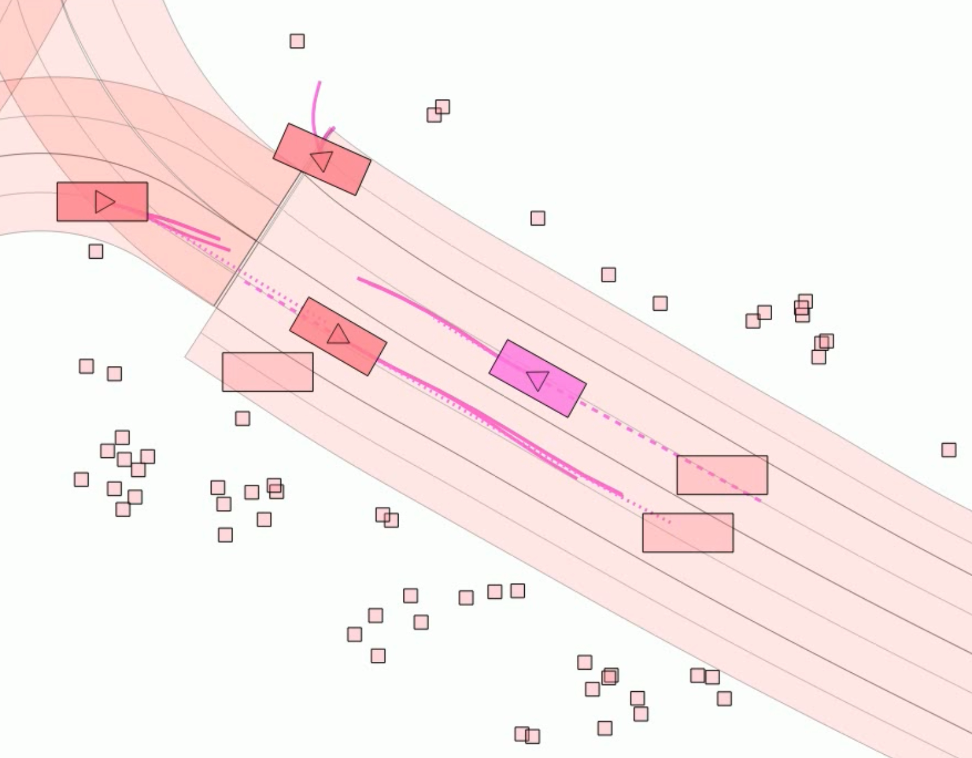} & 
        \includegraphics[width=0.175\linewidth]{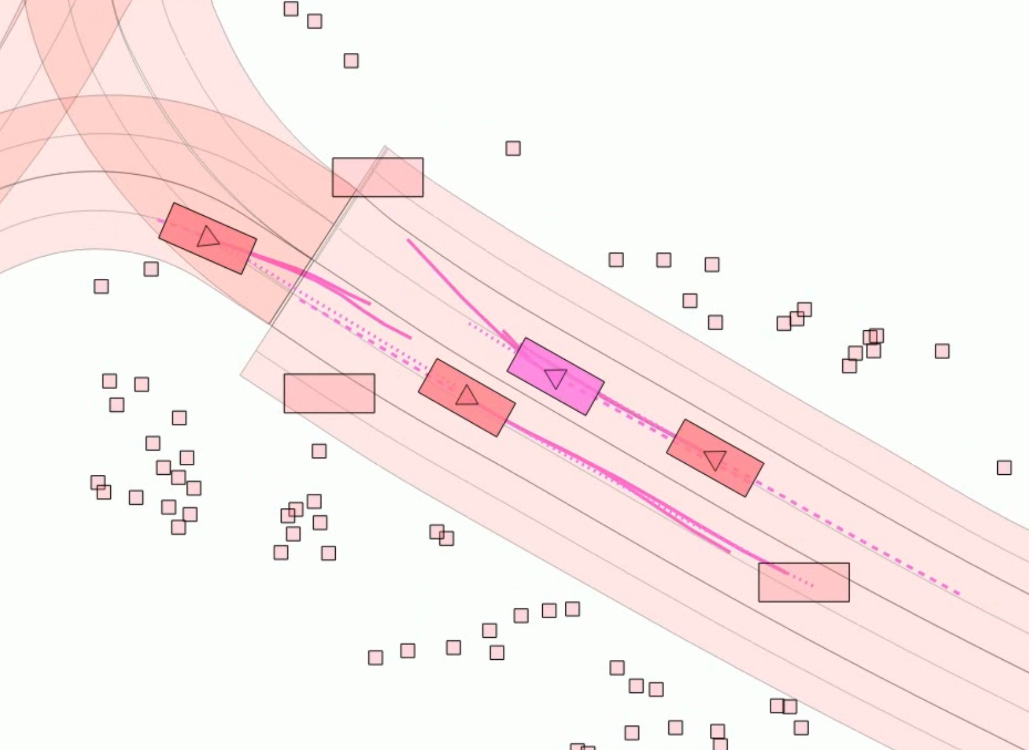} & 
        \includegraphics[width=0.175\linewidth]{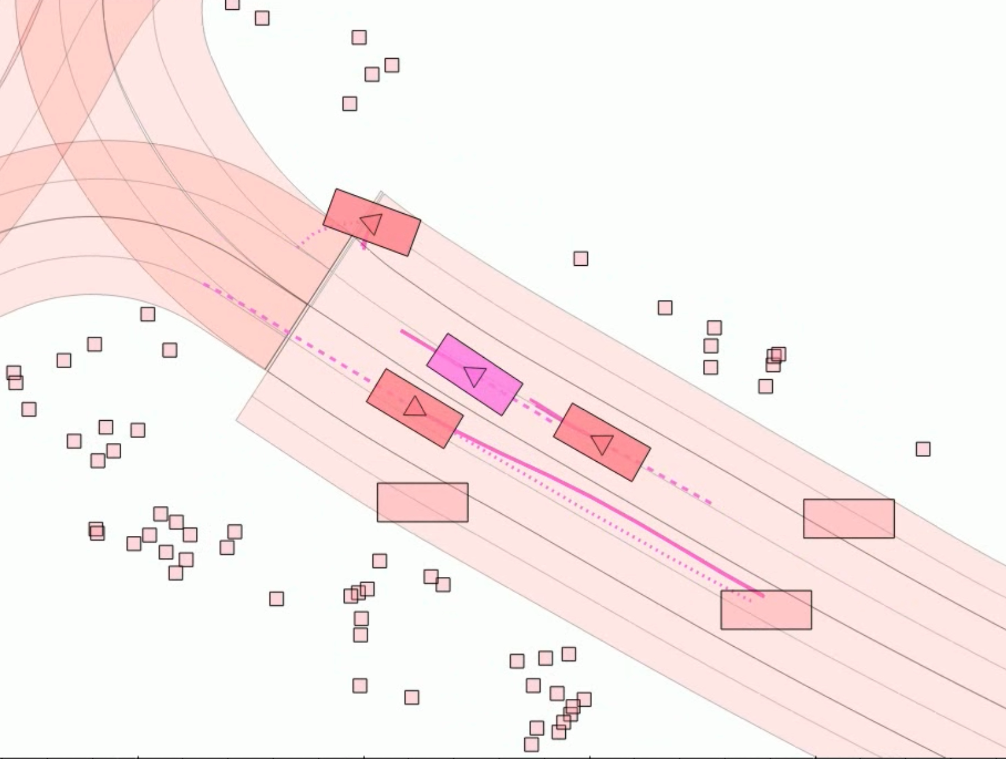} \\
        \includegraphics[width=0.175\linewidth]{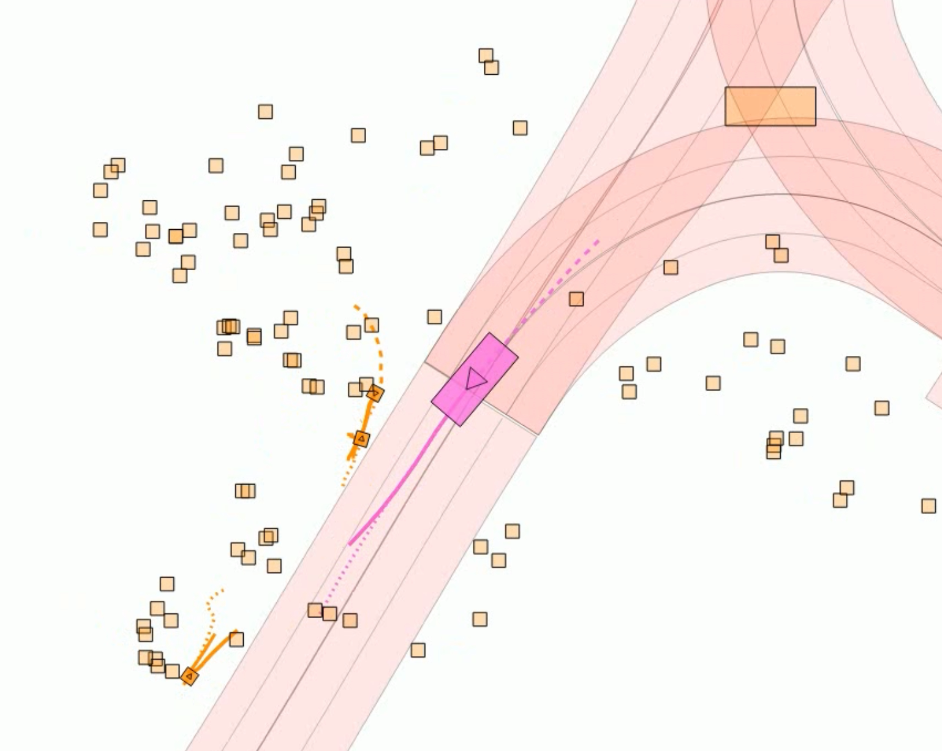} & 
        \includegraphics[width=0.175\linewidth]{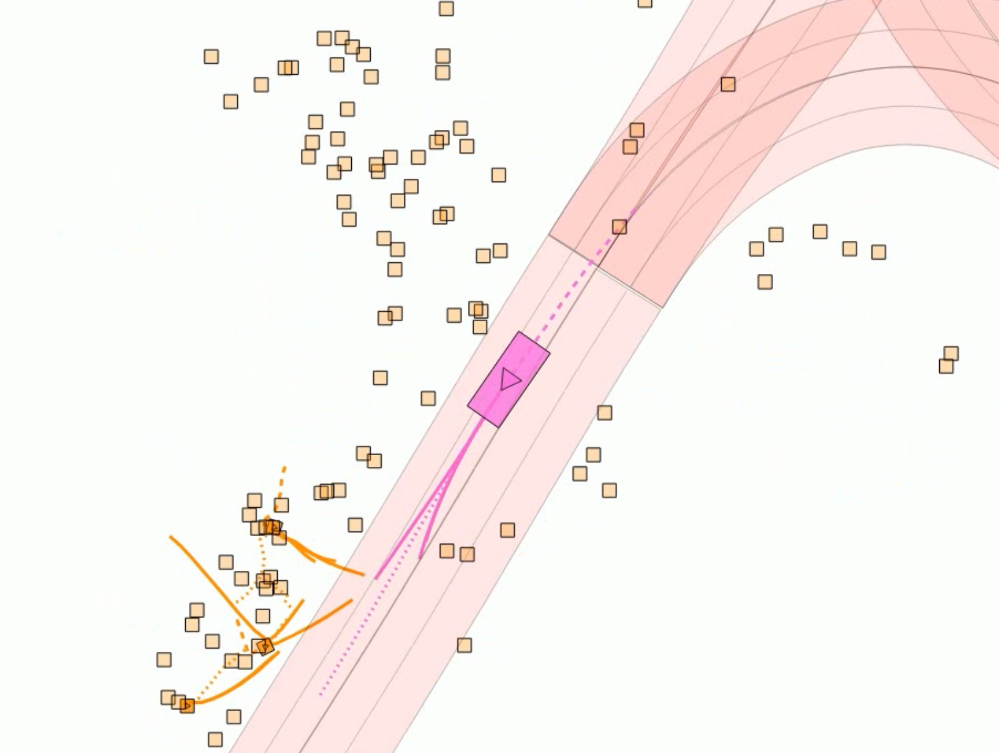} & 
        \includegraphics[width=0.175\linewidth]{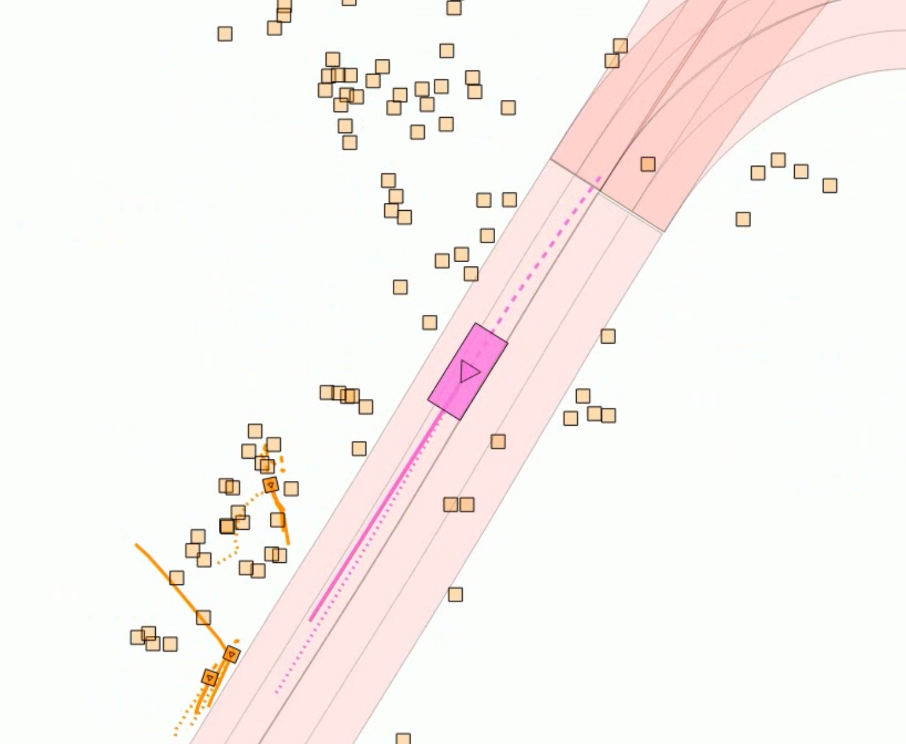} & 
        \includegraphics[width=0.175\linewidth]{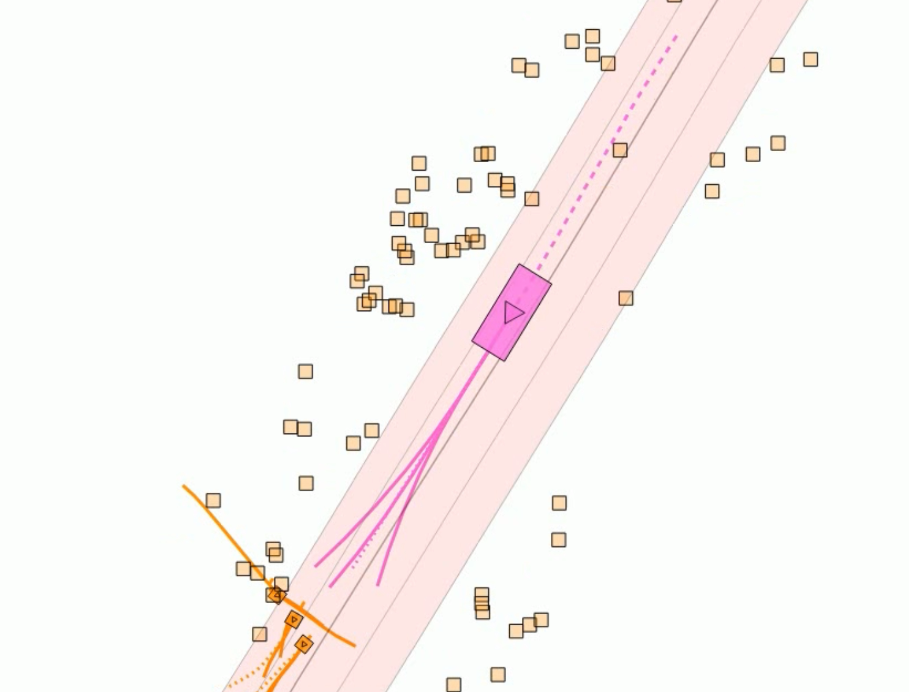} & 
        \includegraphics[width=0.175\linewidth]{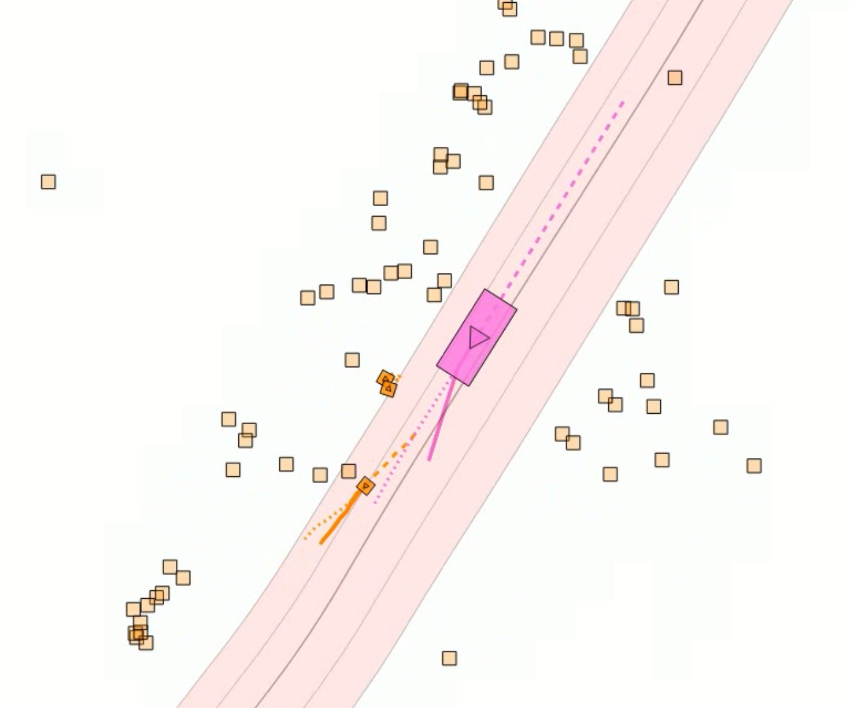} \\
        $t_0$ & $t_0+2.5$ & $t_0+5.0$ & $t_0+7.5$ & $t_0+10.0$
    \end{tabular}
    \caption{Qualitative results on validation scenarios using ScePT: Scenario visualizations show the ego vehicle (pink), pedestrians (orange), and vehicles (red) along with prediction (solid lines), and GT trajectories (dashed lines). We highlight the adaptation of the planner as the ego interacts with accurately predicted pedestrians (first row), precise prediction but with planning errors (second row), dense traffic interactions with slow moving vehicles on the side (third row), and ego-vehicle overtaking, and interacting with pedestrians (last row).}
    \label{fig:baseline}
\end{figure*}

Figure~\ref{fig:baseline} shows four selected complex and dense traffic scenarios at timestep $t_0$, $t_0+2.5s$, $t_0+5.0s$, $t_0+7.5s$, $t_0+10s$, where $t_0$ is chosen as the custom start time for each selected scenario. It can be seen that the DeepUrban data can be used to model realistic and interactive agent prediction and planning trajectories. Especially in scenarios with heavy VRU-Vehicle interaction (first and last row) the baseline shows relevant model characteristics to master the situations.
Interestingly, at $t_0 +7.5s$ (second row), the planned trajectory veers out of the drivable area. This occurs because ScePT's planning implementation relies solely on a reference line—comprising the current lane and its immediate next lane—without integrating the BEV map encoding. Consequently, when multiple next lanes are possible, the first lane (index 0), typically the straight lane, is selected, as seen in this instance.

\begin{table}[h]
    \centering
    \begin{tabularx}{\linewidth}{|X||X|X|X|X|}
        \hline
        & \textbf{Trained: D - Eval: D} & \textbf{Trained: D$_{25}$ - Eval: D} & \textbf{Trained: D$_{50}$ - Eval: D} & \textbf{Trained: D$_{75}$ - Eval: D} \\ \hline
        \textbf{Pred. ADE} & \textbf{0.13} / 0.25 & 0.18 / 0.34 & 0.15 / 0.28 & 0.14 / \textbf{0.25} \\ \hline
        \textbf{Pred. FDE (4s)} & \textbf{0.68} / 1.12 & 0.87 / 1.54 & 0.72 / 1.14 & 0.71 / \textbf{1.11}  \\ \hline
        \textbf{CS \% (4s)} & \textbf{0.072} / 0.299 & 0.087 / \textbf{0.200} & 0.086 / 0.338 & \textbf{0.072} / 0.262 \\ \hline
    \end{tabularx}
    \caption{Table of ScePT vehicle~/~pedestrian best-of-1 Baseline Validation \small{(D = Munich Tal trained on 100\%, D$_{25}$ = Munich Tal trained on 25\%, D$_{50}$ = Munich Tal trained on 50\%, D$_{75}$ = Munich Tal trained on 75\%)}}
    \label{tab:baseline}
\end{table}

\mycomment{
\begin{table}[h]
    \centering
    \begin{tabularx}{\linewidth}{|X||X|X|X|X|}
        \hline
        & \textbf{Trained: D - Eval: D} & \textbf{Trained: D$_{25}$ - Eval: D} & \textbf{Trained: D$_{50}$ - Eval: D} & \textbf{Trained: D$_{75}$ - Eval: D} \\ \hline
        \textbf{Pred. ADE} & 0.12 / 0.23 & \textbf{0.11} / \textbf{0.20} & 0.12 / 0.22 & 0.13 / 0.22 \\ \hline
        \textbf{Pred. FDE (4s)} & 0.62 / 1.01 & 0.54 / \textbf{0.85} & \textbf{0.38} / 0.98 & 0.64 / 0.94  \\ \hline
        \textbf{CS \% (4s)} & \textbf{0.075} / \textbf{0.252} & 0.101 / 0.256 & 0.094 / 0.309 & \textbf{0.075} / 0.256 \\ \hline
    \end{tabularx}
    \caption{Table of ScePT vehicle~/~pedestrian best-of-3 Baseline Validation \small{(D = Munich Tal trained on 100\%, D$_{25}$ = Munich Tal trained on 25\%, D$_{50}$ = Munich Tal trained on 50\%, D$_{75}$ = Munich Tal trained on 75\%)}}
    \label{tab:baseline}
\end{table}
}

\subsection{Generalization capabilities}
To assess the generalization performance of the DeepUrban dataset we perform cross-validation on various subsets of the dataset and the nuScenes dataset.
\subsubsection{Cross-Area}
Table~\ref{tab:crossarea} provides details on the training and validation datasets employed across the four locations in DeepUrban, offering insights into the generalization capabilities over various terrains and scenarios. The Table shows that the usage of additional locations can further improve results, concerning the metrics ADE and FDE. Not surprisingly, performance benefits from adding area-specific data (e.g. US locations when evaluating on US data).

\begin{table*}[h]
    \centering
    \begin{tabularx}{\textwidth}{|X||X||X|X|X||X|X|X|}
        \hline
        & \textbf{Trained: D - Eval: D} & \textbf{Trained: D - Eval: D$^+$} & \textbf{Trained: D$^+$ - Eval: D$^+$} & \textbf{Trained: D$^{+US}$ - Eval: D$^+$} & \textbf{Trained: D - Eval: D$^{US}$} & \textbf{Trained: D$^+$ - Eval: D$^{US}$} & \textbf{Trained: D$^{+US}$ - Eval: D$^{US}$} \\ \hline
        \textbf{Pred. ADE} & 0.13 / 0.25 & 0.22 / 0.25 & \textbf{0.19} / \textbf{0.24} & \textbf{0.19} / \textbf{0.24} & 0.15 / 0.09 & 0.18 / 0.09 & \textbf{0.11} / \textbf{0.08} \\ \hline
        \textbf{Pred. FDE (4s)} & 0.68 / 1.12 & 1.14 / 1.11 & \textbf{0.98} / 1.08 & \textbf{0.98} / \textbf{1.05} & 0.80 / 0.41 & 0.98 / 0.41 & \textbf{0.60} / \textbf{0.38} \\ \hline
        \textbf{CS \% (4s)} & 0.0722 / 0.2989 & 0.0655 / \textbf{0.2924} & \textbf{0.0417} / 0.3049 & 0.0442 / 0.3717 & 0.0074 / \textbf{0.0357} & 0.0086 / 0.0565 & \textbf{0.0025} / 0.0713 \\ \hline
    \end{tabularx}
    \caption{Cross-validation on various DeepUrban locations showing vehicle~/~pedestrian ADE, FDE and collision score (CS) for prediction horizons of 4s. DeepUrban locations: D = Munich Tal only, D$^+$ = all German locations, D$^{+US}$ = all German locations + US, D$^{US}$ = US only}
    \label{tab:crossarea}
\end{table*}

\mycomment{
\begin{table*}[h]
    \centering
    \begin{tabularx}{\textwidth}{|X||X||X|X|X||X|X|X|}
        \hline
        & \textbf{Trained: D - Eval: D} & \textbf{Trained: D - Eval: D$^+$} & \textbf{Trained: D$^+$ - Eval: D$^+$} & \textbf{Trained: D$^{+US}$ - Eval: D$^+$} & \textbf{Trained: D - Eval: D$^{US}$} & \textbf{Trained: D$^+$ - Eval: D$^{US}$} & \textbf{Trained: D$^{+US}$ - Eval: D$^{US}$} \\ \hline
        \textbf{Pred. ADE} & 0.12 / 0.23 & \textbf{0.12} / 0.24 & \textbf{0.12} / 0.24 & 0.13 / \textbf{0.14} & \textbf{0.08} / 0.09 & 0.12 / \textbf{0.08} & 0.09 / \textbf{0.08} \\ \hline
        \textbf{Pred. FDE (4s)} & 0.62 / 1.01 & 0.62 / 1.01 & \textbf{0.61} / 1.02 & 0.63 / \textbf{0.99} & 0.73 / 0.37 & 0.64 / 0.37 & \textbf{0.44} / \textbf{0.33} \\ \hline
        \textbf{CS \% (4s)} & 0.0746 / 0.2524 & 0.0739 / \textbf{0.2519} & 0.0721 / 0.3065 & \textbf{0.0442} / 0.3717 & 0.0110 / \textbf{0.0147} & 0.0089 / 0.0545 & \textbf{0.0067} / 0.0603 \\ \hline
    \end{tabularx}
    \caption{Table of ScePT for Cross Area vehicle/pedestrian best-of-3 Validation on DeepUrban(D) \small{(D = Munich Tal only, D$^+$ = all German, D$^{+US}$ = all German + US, D$^{US}$ = US only)}}
    \label{tab:crossarea}
\end{table*}
}

\subsubsection{Cross Dataset}
\begin{figure*}[ht]
    \begin{subfigure}[b]{\linewidth}
        \centering
        \begin{subfigure}[b]{0.24\linewidth}
            \includegraphics[width=\linewidth]{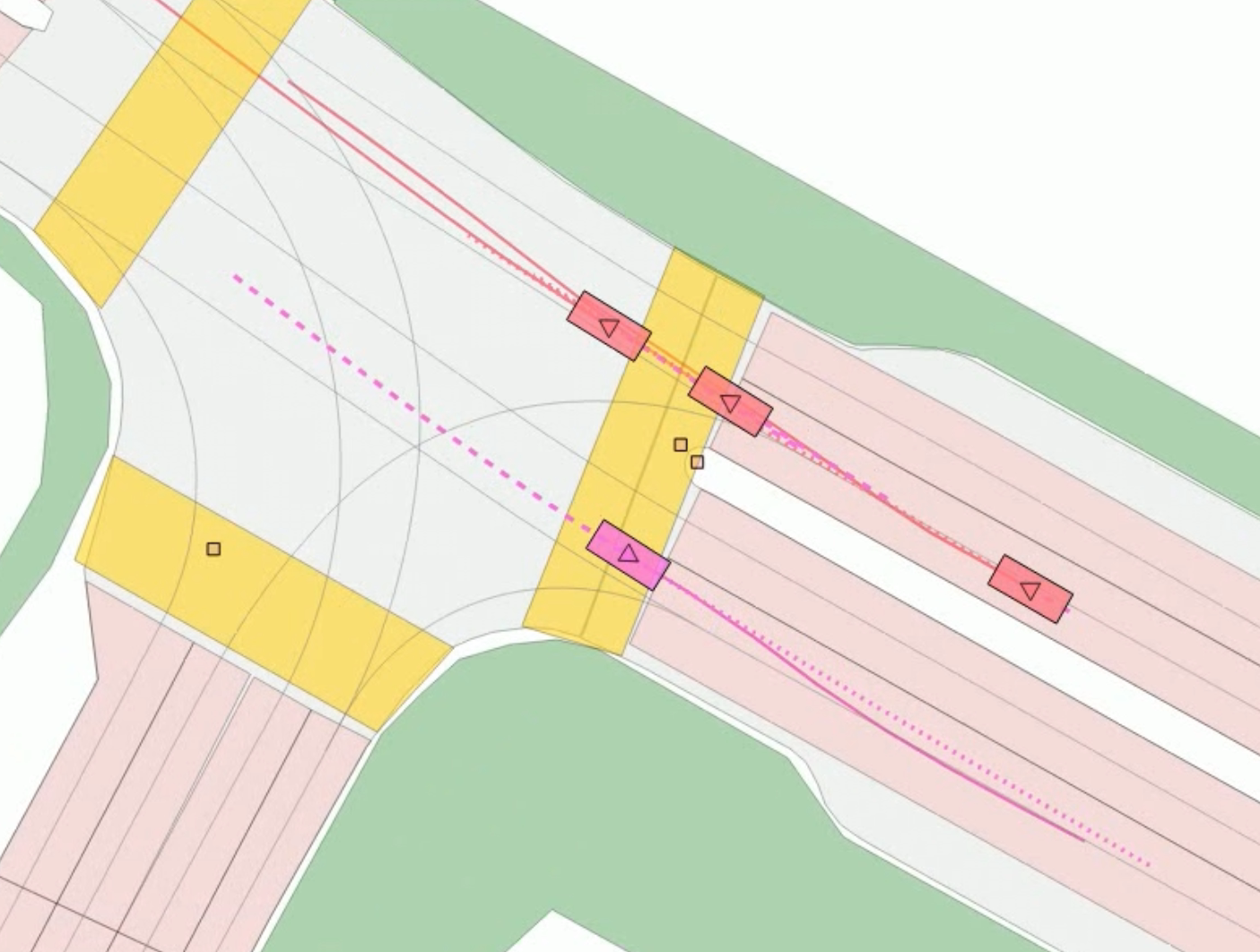}
        \end{subfigure}
        \begin{subfigure}[b]{0.24\linewidth}
            \includegraphics[width=\linewidth]{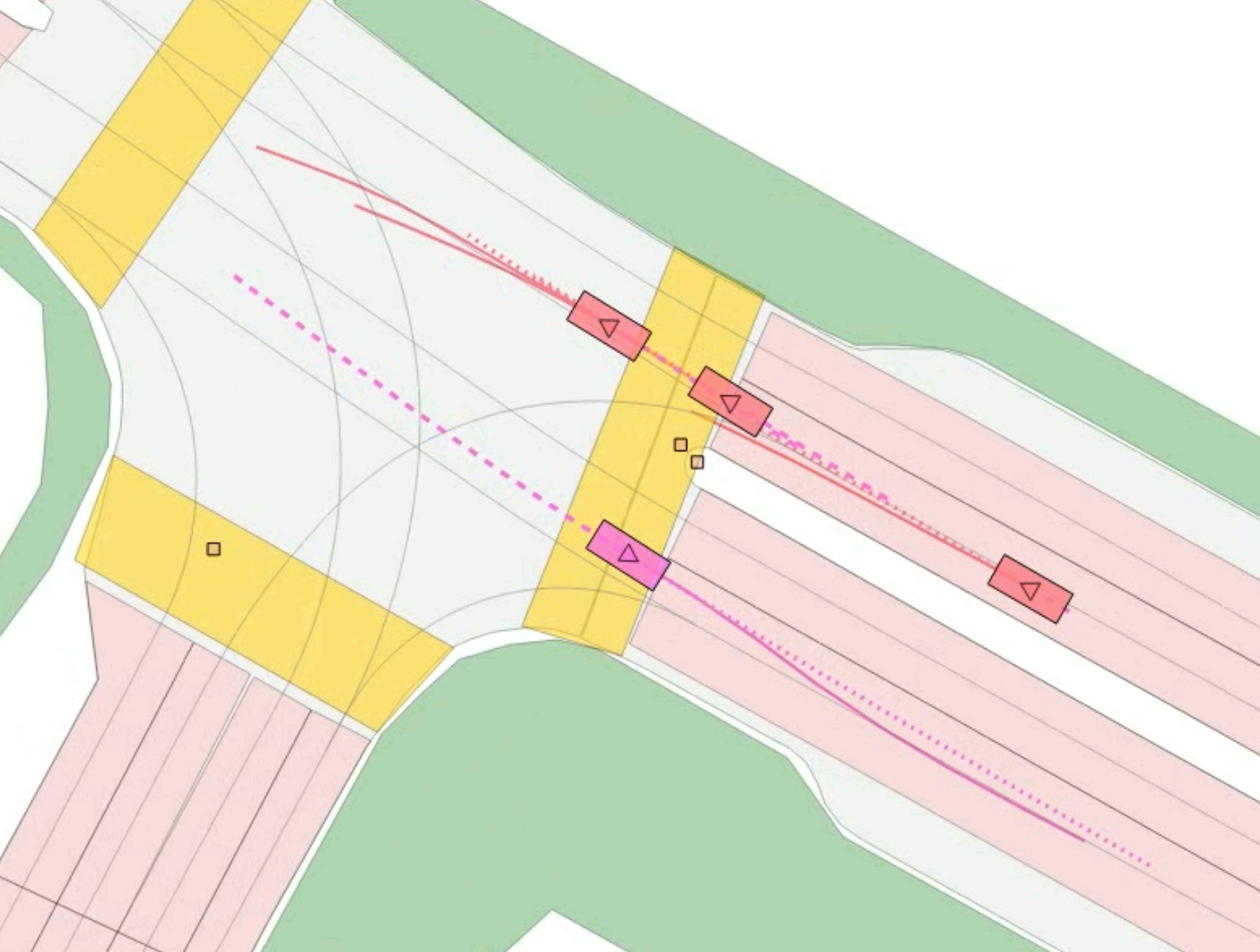}
        \end{subfigure}
        \hfill
        \begin{subfigure}[b]{0.24\linewidth}
            \includegraphics[width=\linewidth]{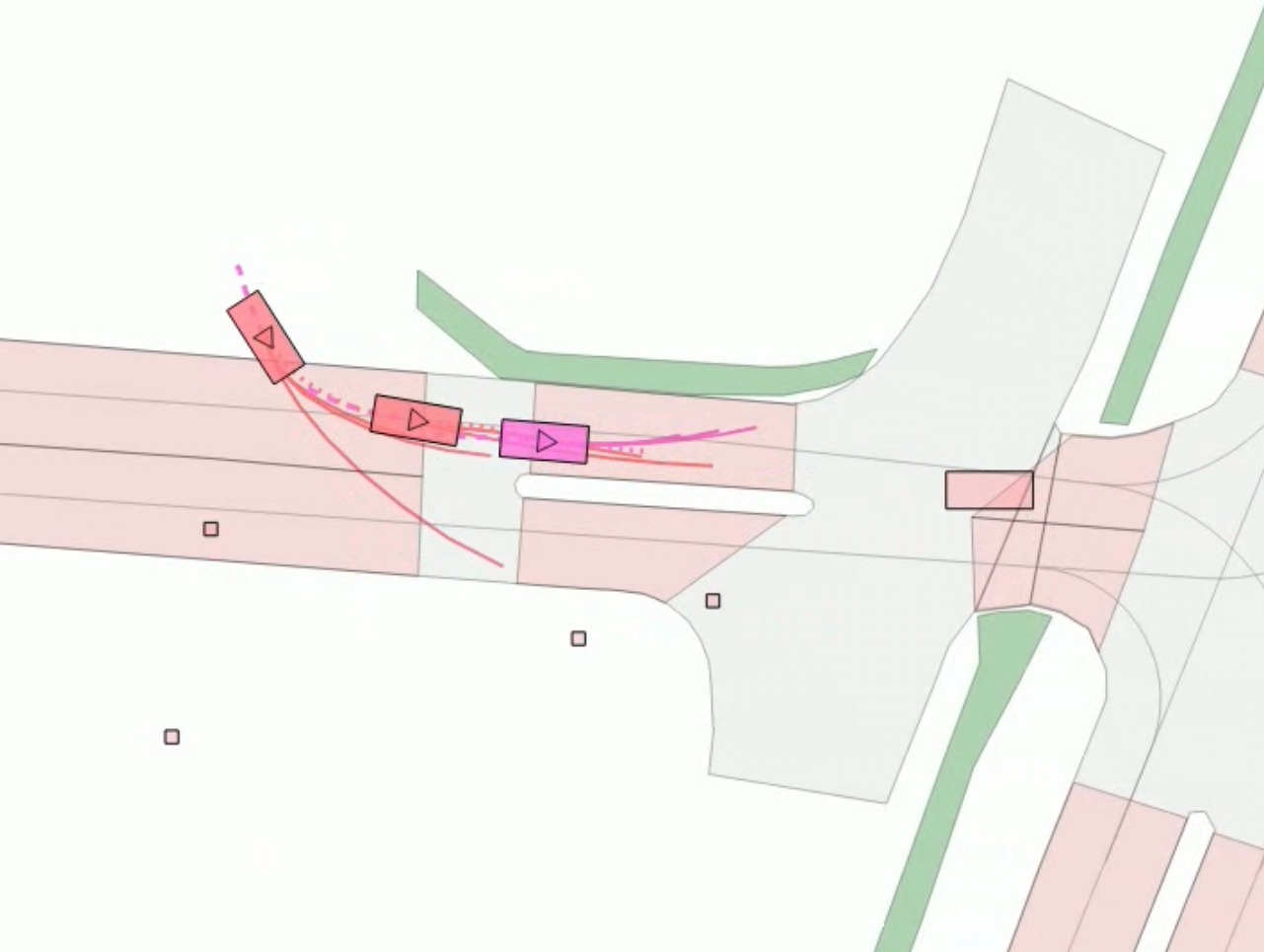}
        \end{subfigure}
        \begin{subfigure}[b]{0.24\linewidth}
            \includegraphics[width=\linewidth]{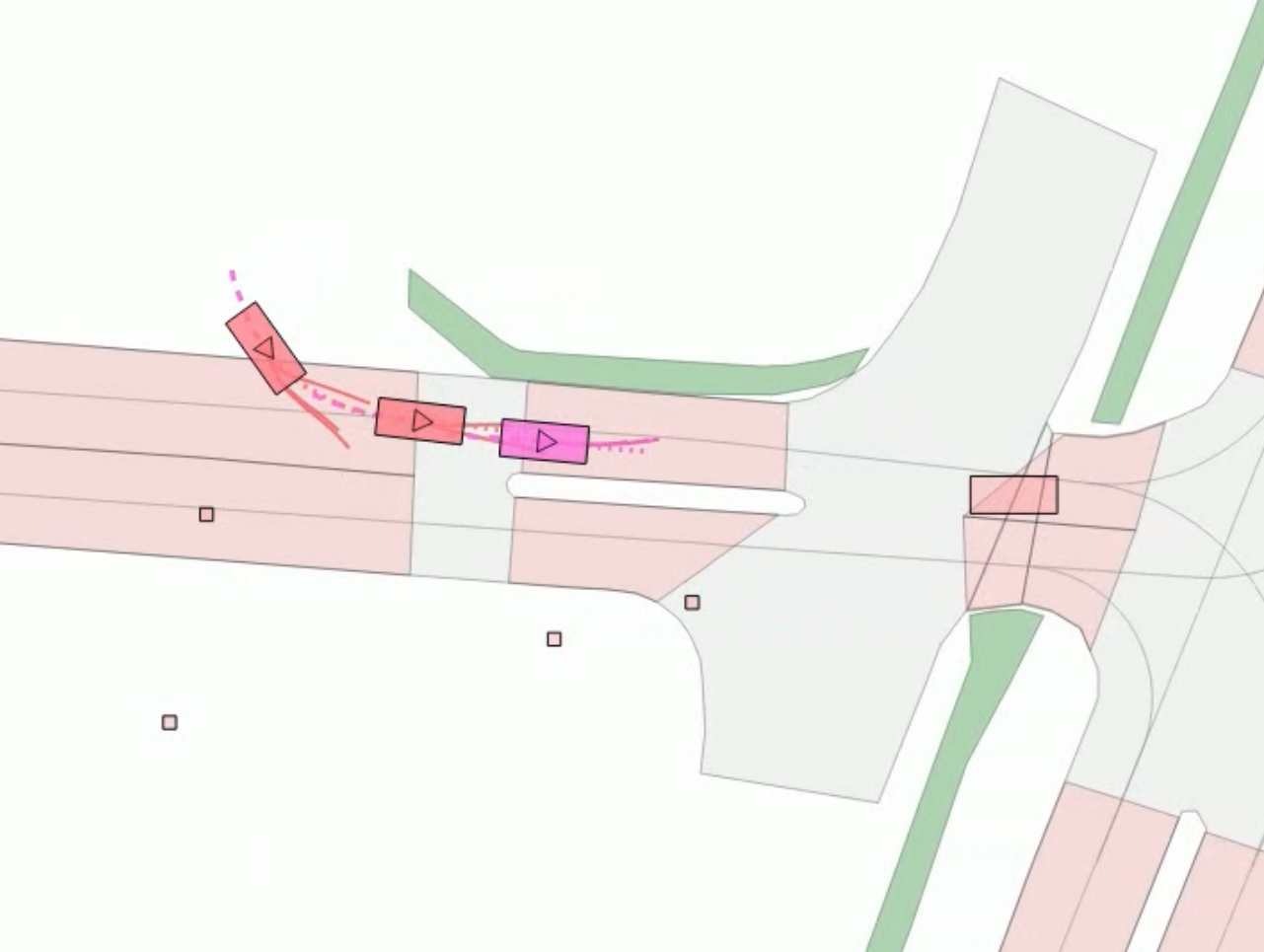}
        \end{subfigure}
    \end{subfigure}
    \begin{subfigure}[b]{\linewidth}
        \centering
        \begin{subfigure}[b]{0.24\linewidth}
            \includegraphics[width=\linewidth]{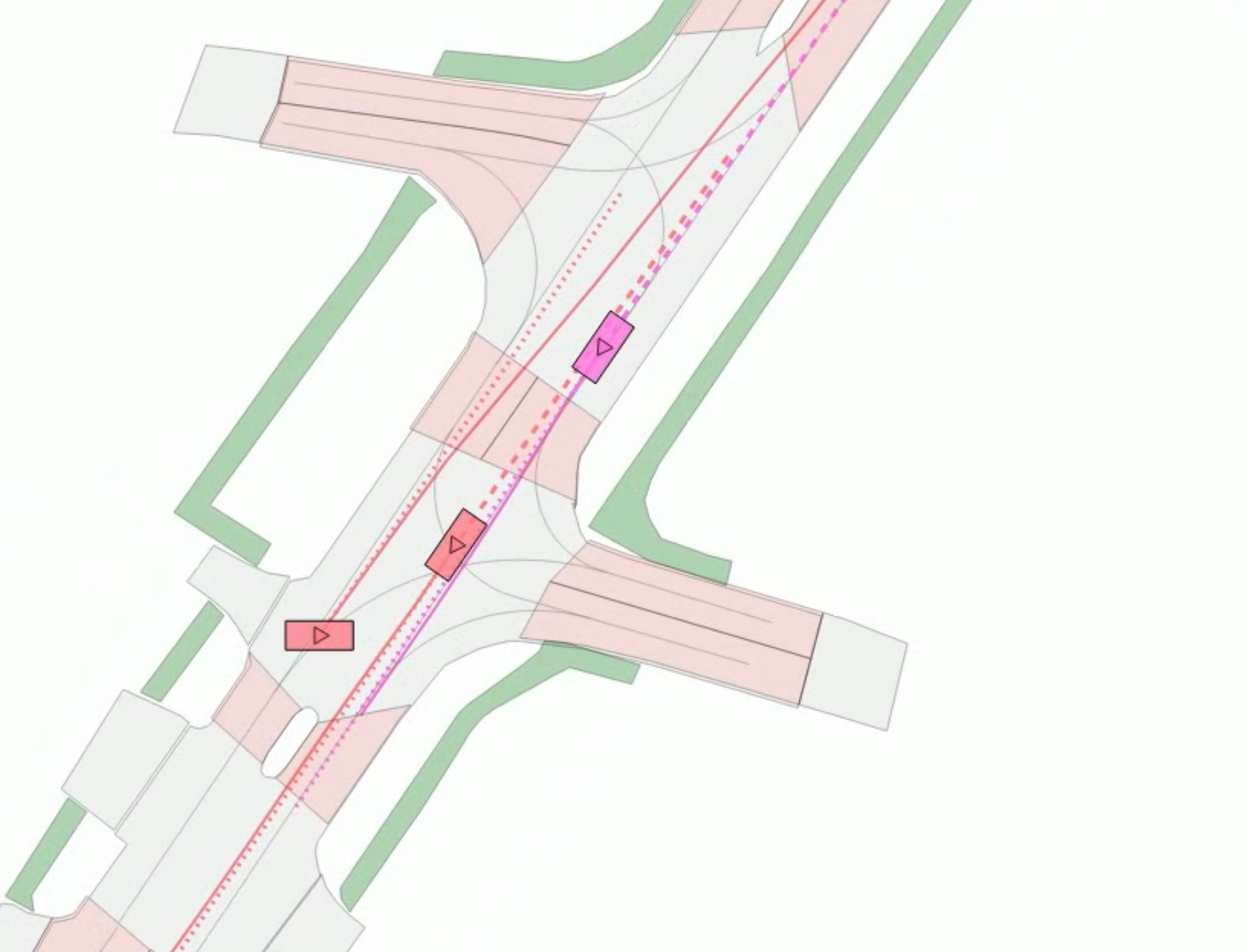}
        \end{subfigure}
        \begin{subfigure}[b]{0.24\linewidth}
            \includegraphics[width=\linewidth]{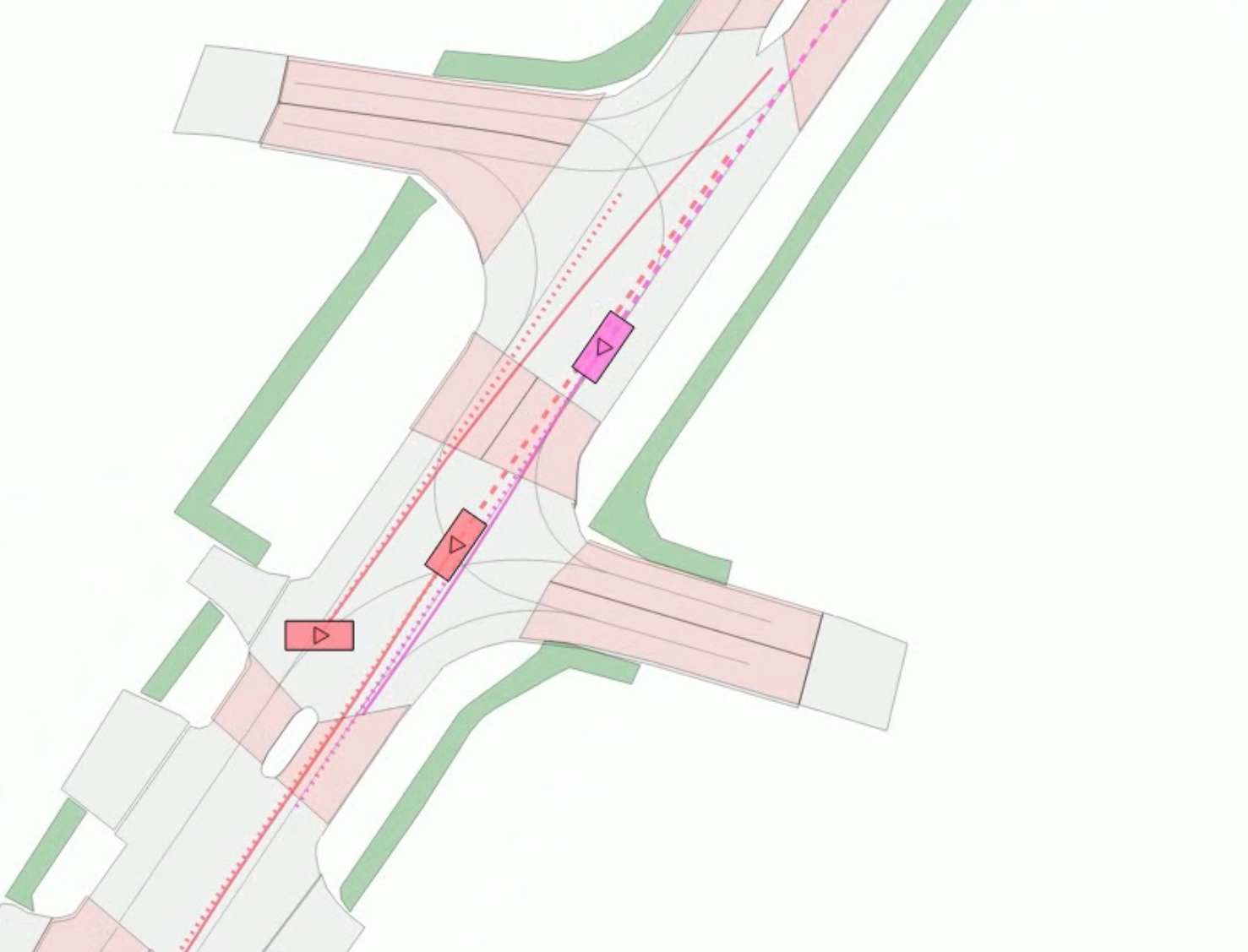}
        \end{subfigure}
        \hfill
        \begin{subfigure}[b]{0.24\linewidth}
            \includegraphics[width=\linewidth]{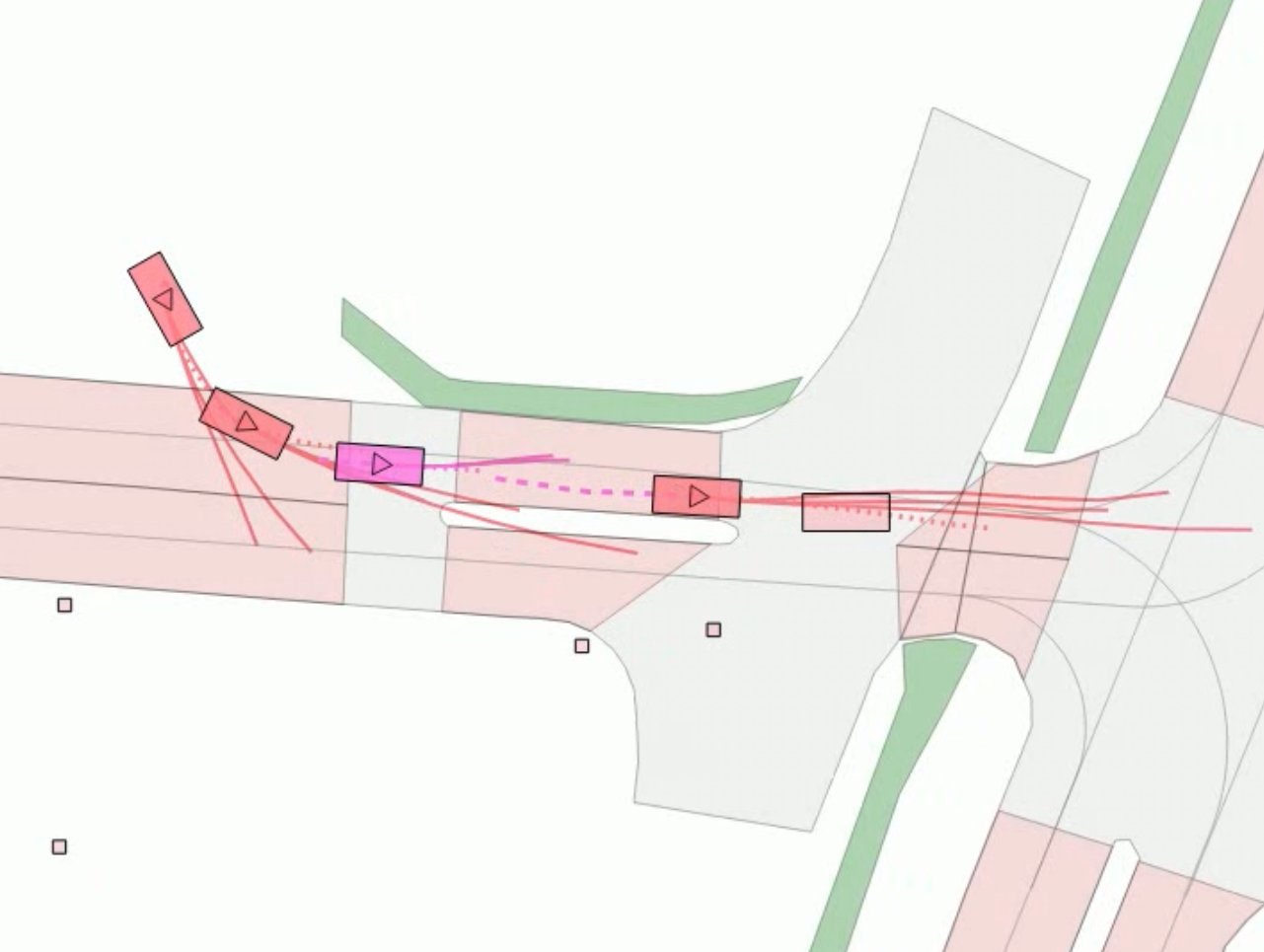}
        \end{subfigure}
        \begin{subfigure}[b]{0.24\linewidth}
            \includegraphics[width=\linewidth]{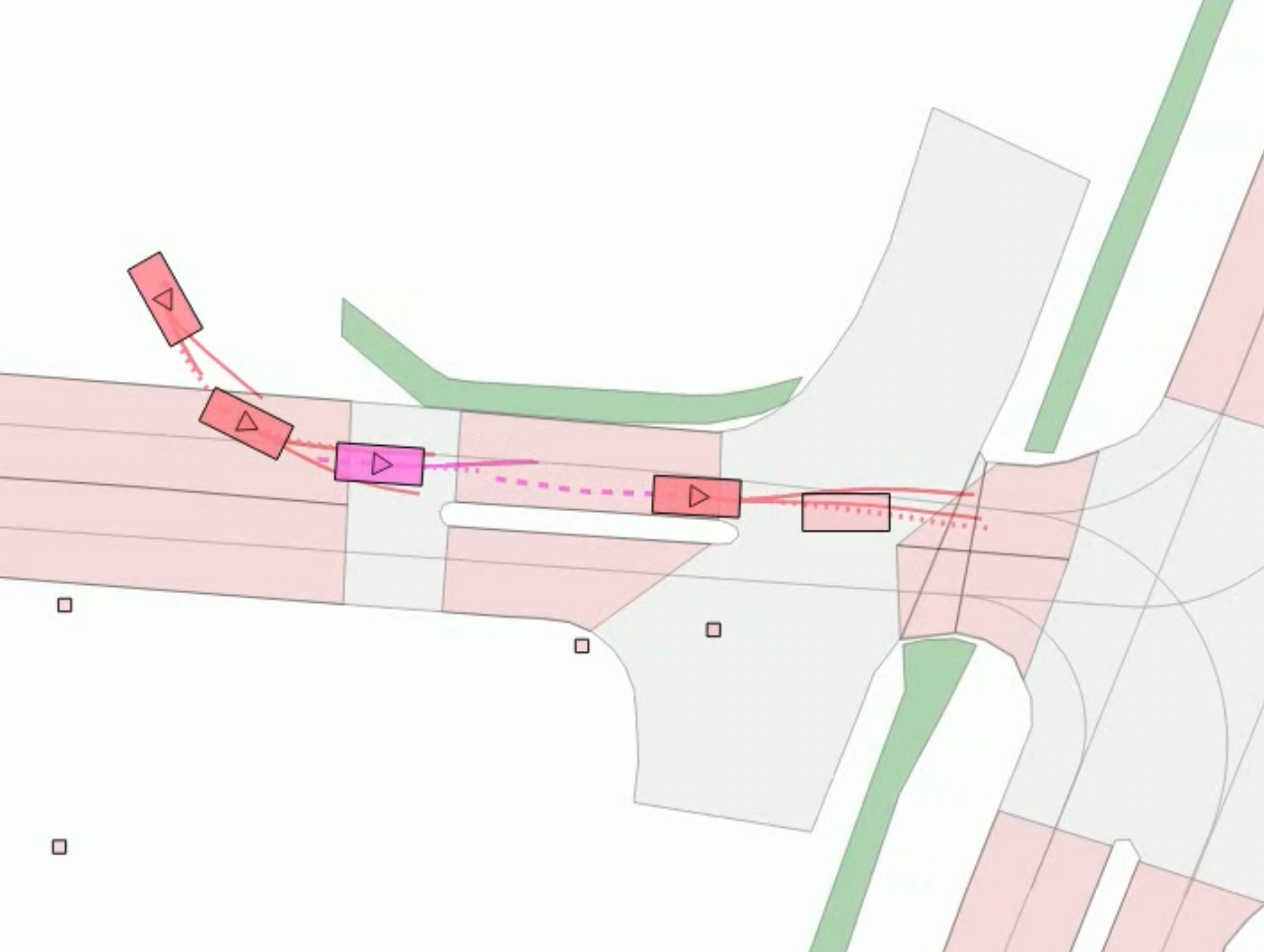}
        \end{subfigure}
    \end{subfigure}
    \caption{Four scenarios validated on nuScenes: Trained on nuScenes(N) (left) and trained on nuScenes + DeepUrban (ND) (right). Scenario visualizations show the ego vehicle (pink), pedestrians (orange), and vehicles (red) along with prediction (solid lines), and GT trajectories (dashed lines). 
    We show: Biggest differences in ADE (here also biggest in FDE) with one mode / three modes (upper left pair / upper right pair). When trained on nuScenes only, vehicle predictions seem to be less conservative, resulting in bigger FDE and ADE;  
    Biggest difference in collision score with one mode / three modes (lower left pair / lower right pair).
    When trained on nuScenes and DeepUrban, predictions seem to be more passive resulting in fewer collisions.}
    \label{fig:crossdataset}
\end{figure*}

We cross-validate our baseline method on nuScenes. Here we use the original nuScenes scenario splits of (500 train~/~150 val~/~150 test) with each scenario lasting 20 seconds. When using the custom ScePT data split\cite{Trajectron2020}, results may differ slightly. Cross-dataset validation results are shown in Table~\ref{tab:crossdataset}. Figure~\ref{fig:crossdataset} shows scenario examples automatically selected based on the biggest metric differences between models trained on nuScenes (N) and nuScenes + DeepUrban Munich Tal (ND). The results demonstrate both qualitatively and quantitatively that incorporating DeepUrban scenarios to the training dataset of nuScenes can enhance prediction and planning performance up to 44.1\% / 44.3\% on the ADE / FDE metrics. 

\begin{table*}[h]
    \vspace{0.2cm}
    \centering
    \begin{tabularx}{\linewidth}{|X||X||X|X|X|X|X|}
        \hline
         & \textbf{Trained: D - Eval: D} & \textbf{Trained: N - Eval: N} & \textbf{Trained: D - Eval: N} & \textbf{Trained: ND - Eval: N} & \textbf{Trained: ND$^+$ - Eval: N} & \textbf{Trained: ND$^{+US}$ - Eval: N} \\ \hline
        \textbf{Pred. ADE} & 0.13 / 0.25 & 0.59 / 0.22 & 0.60 / 0.21 & \textbf{0.33} / 0.20 & 0.34 / 0.19  & 0.34 / \textbf{0.19} \\ \hline
        \textbf{Pred. FDE (4s)} & 0.68 / 1.12 & 3.00 / 1.04 & 3.05 / 0.99 & \textbf{1.67} / 0.98 & 1.73 / \textbf{0.93} & 1.72 / \textbf{0.93} \\ \hline
        \textbf{CS \% (4s)} & 0.0722 / 0.2989 & 0.0507 / \textbf{0.1161} & 0.0545 / 0.1578 & 0.0254 / 0.1841 & \textbf{0.0198} / 0.1797 & 0.0259 / 0.2170 \\ \hline
    \end{tabularx}
    \caption{Cross-validation on nuScenes (N) and DeepUrban (D) showing vehicle~/~pedestrian ADE, FDE and collision score (CS) for prediction horizons of 4s. DeepUrban locations: D = Munich Tal only, D$^+$ = all German locations, D$^{+US}$ = all German locations + US, D$^{US}$ = US only}
    \label{tab:crossdataset}
\end{table*}

\mycomment{
\begin{table*}[h]
    \centering
    \begin{tabularx}{\linewidth}{|X||X||X|X|X|X|X|}
        \hline
         & \textbf{Trained: D - Eval: D} & \textbf{Trained: N - Eval: N} & \textbf{Trained: D - Eval: N} & \textbf{Trained: ND - Eval: N} & \textbf{Trained: ND$^+$ - Eval: N} & \textbf{Trained: ND$^{+US}$ - Eval: N} \\ \hline
        \textbf{Pred. ADE} & 0.12 / 0.23 & \textbf{0.24} / \textbf{0.16} & 0.57 / 0.19 & 0.29 / 0.19 & 0.29 / 0.19  & 0.30 / 0.18 \\ \hline
        \textbf{Pred. FDE (4s)} & 0.62 / 1.01 & \textbf{1.13} / \textbf{0.77} & 2.92 / 0.91 & 1.44 / 0.90 & 1.43 / 0.90 & 1.48 / 0.83 \\ \hline
        \textbf{CS \% (4s)} & 0.0746 / 0.2524 & 0.0431 / \textbf{0.1225} & 0.0565 / 0.1313 & 0.0258 / 0.1812 & 0.0258 / 0.1812 & \textbf{0.0266} / 0.1929 \\ \hline
    \end{tabularx}
    \caption{Table of ScePT for Cross Dataset vehicle/pedestrian best-of-3 Validation on nuScenes(N) and DeepUrban(D)
    \small{(D = Munich Tal only, D$^+$ = all German, D$^{+US}$ = all German + US)}}
    \label{tab:crossdataset}
\end{table*}
}

\section{Discussion}
Our results suggest the usability of DeepUrban. Quantitative and qualitative results suggest that state-of-the-art methods can be trained on DeepUrban and capture complex dense urban traffic situations. Training on DeepUrban scenarios alongside nuScenes has improved the performance when evaluated on nuScenes. Incorporating DeepUrban leads to more stable and passive trajectories, as evidenced by the qualitative results in Figure~\ref{fig:crossdataset}.
For comparison purposes, it was advantageous to use SecPT with its recommended clique size of four agents for the nuScenes dataset, significantly limiting the number of influencing agents. In high-density areas like 'Munich Tal', where there can be up to 50 agents within a 20-meter radius, algorithms with explicit interaction modelling will face substantial compute challenges. Further the high-density area itself results in more complex scenarios due to the increased number of agents, leading to more passive or even halted ego-motion. To utilize drone data particularly for planning, e.g. modelling the visibility around the vehicle for each timestep may be necessary, as not all agents within a radius are visible at all times.
We note here, that evaluating the collision score for pedestrians (according to \cite{ScePT2022}) may be less reliable due to imprecise static size boundary boxes in crowded settings and pedestrian dynamics. Future research will further explore this generalization through cross-dataset validation on other datasets like nuPlan, Waymo, etc. 
While we are still recording new locations with our partner DeepScenario, we will gradually add and test on more data in the near future.
To facilitate objective benchmarking on the DeepUrban dataset, we will release the dataset and extensions along with an online evaluation server on our website (\href{https://iv.ee.hm.edu/deepurban}{https://iv.ee.hm.edu/deepurban}).

\section{Conclusion}
In this work we introduced the DeepUrban dataset, developed in collaboration with DeepScenario. We showed that DeepUrban improves trajectory prediction and planning for autonomous driving in dense urban environments. By incorporating detailed 3D traffic objects and comprehensive environmental data we showed that DeepUrban is appropriate for this task. Furthermore, we showed that when expanding existing datasets with DeepUrban, we achieved improvements up to 44.1\% / 44.3\% on the ADE / FDE metrics. This suggests that the DeepUrban dataset can help to improve the reliability and effectiveness of autonomous driving technologies in complex traffic scenarios.

\section{Future Work}
We are planning significant expansions to our prediction and planning benchmark to enhance its utility and robustness. Firstly, we intend to develop an Evaluation Server. This server will be dedicated to testing algorithms on scenarios that were not included in the initial proposal. The goal is to assess how these algorithms perform under a hidden test set.  Additionally, the dataset will be extended by modelling the visibility and incorporating more recorded locations across Germany.

Secondly, we plan to integrate our benchmark into Carla, a renowned simulation platform, to provide a rigorous testing environment that mimics real-world complexities. This will enable detailed evaluation of planning algorithms in dynamic scenarios, bridging the gap between theoretical advancements and practical applications in autonomous driving technologies.

\section*{ACKNOWLEDGMENT}
This work is a result of the joint research project STADT:up (19A22006N). The project is supported by the German Federal Ministry for Economic Affairs and Climate Action (BMWK), based on a decision of the German Bundestag. The author is solely responsible for the content of this publication. 

Further, the authors gratefully acknowledge the support of DeepScenario GmbH for providing the essential data for this research work.

\end{document}